\documentclass{article}

\usepackage{microtype}
\usepackage{graphicx}
\usepackage{subfigure}
\usepackage{booktabs} % for professional tables
\usepackage{inconsolata}
\usepackage{amsfonts}
\usepackage{multirow}
\usepackage{subcaption}
\usepackage{hyperref}

\usepackage[accepted]{icml2025}

\usepackage{amsmath}
\usepackage{amssymb}
\usepackage{mathtools}
\usepackage{amsthm}
\usepackage{amsmath,amsfonts,bm}

\def\eqref#1{equation~\ref{#1}}
\def\1{\bm{1}}

\def\vtheta{{\bm{\theta}}}

\def\vd{{\bm{d}}}

\def\vh{{\bm{h}}}

\def\vo{{\bm{o}}}
\def\vp{{\bm{p}}}

\def\vu{{\bm{u}}}
\def\vv{{\bm{v}}}
\def\vw{{\bm{w}}}
\def\vx{{\bm{x}}}

\def\vz{{\bm{z}}}

\def\mA{{\bm{A}}}

\def\mC{{\bm{C}}}

\def\mE{{\bm{E}}}

\def\mM{{\bm{M}}}

\def\mO{{\bm{O}}}

\def\mS{{\bm{S}}}

\def\mV{{\bm{V}}}

\DeclareMathAlphabet{\mathsfit}{\encodingdefault}{\sfdefault}{m}{sl}
\SetMathAlphabet{\mathsfit}{bold}{\encodingdefault}{\sfdefault}{bx}{n}

\newcommand{\R}{\mathbb{R}}

\usepackage{xcolor}
\definecolor{citecolor}{HTML}{2779af}
\definecolor{linkcolor}{HTML}{c0392b}
\usepackage{graphicx}
\usepackage{tablefootnote}
\usepackage[group-separator={,},group-minimum-digits=3,reset-text-series=false,text-series-to-math=true,unit-mode=text]{siunitx}
\usepackage{enumitem}
\usepackage{verbatim}

\usepackage{tabularx}
\usepackage{algorithm}
\usepackage[indLines=true,noEnd=false,italicComments=false,commentColor=black]{algpseudocodex}

\usepackage{array,etoolbox}
\preto\tabular{\setcounter{magicrownumbers}{0}}
\newcounter{magicrownumbers}
\def\rownumber{}

\usepackage[capitalize,noabbrev]{cleveref}

\def\vtheta{\bm{\theta}}
\def\vxigma{\bm{\sigma}}
\def\valpha{\bm{\alpha}}
\def\vell{\bm{\ell}}
\def\mLSE{\textcolor{black}{\mathrm{LSE}}}
\def\mone{\textcolor{blue}{(1)}}
\def\mtwo{\textcolor{blue}{(2)}}
\def\mthree{\textcolor{blue}{(3)}}

\newcommand{\lrbrackround}[1]{\left(#1\right)}

\newcommand{\lrbrackvec}[1]{\left\|#1\right\|}
\newcommand{\lrbracknorm}[1]{\left|#1\right|}

\theoremstyle{plain}
\newtheorem{theorem}{Theorem}[section]
\newtheorem{proposition}[theorem]{Proposition}
\newtheorem{lemma}[theorem]{Lemma}

\theoremstyle{definition}
\newtheorem{definition}[theorem]{Definition}

\theoremstyle{remark}
\newtheorem{remark}[theorem]{Remark}

\usepackage[textsize=tiny]{todonotes}

\icmltitlerunning{Calibrated Large Language Models and How to Find Them with Label Smoothing}

\begin{document}

\twocolumn[
\icmltitle{Calibrated Language Models and How to Find Them with Label Smoothing}



\icmlsetsymbol{equal}{$\sharp$}
% \icmlsetsymbol{senior}{$\flat$}

\begin{icmlauthorlist}
\icmlauthor{Jerry Huang}{equal,udem,mila}
\icmlauthor{Peng Lu}{equal,udem}
\icmlauthor{Qiuhao Zeng}{uwo}
\end{icmlauthorlist}

\icmlaffiliation{udem}{Universit\'{e} de Montr\'{e}al}
\icmlaffiliation{mila}{Mila - Quebec AI Institute}
\icmlaffiliation{uwo}{University of Western Ontario}
% \icmlaffiliation{nal}{Noah's Ark Lab}

\icmlcorrespondingauthor{Jerry Huang}{\href{mailto:jerry.huang@mila.quebec}{\texttt{jerry.huang@mila.quebec}}}
\icmlcorrespondingauthor{Peng Lu}{\href{mailto:peng.lu@umontreal.ca}{\texttt{peng.lu@umontreal.ca}}}
% \icmlcorrespondingauthor{Qiuhao Zeng}{\href{mailto:qzeng53@uwo.ca}{\texttt{qzeng53@uwo.ca}}}

\icmlkeywords{
    ICML, 
    Machine Learning, 
    Natural Language Processing, 
    Confidence Calibration,
    Large Language Models,
    Label Smoothing
}
\vskip 0.3in
]



\printAffiliationsAndNotice{\icmlEqualContribution} % otherwise use the standard text.

\begin{abstract}
Recent advances in natural language processing have enabled the fine-tuning of large language models (LLMs) into powerful interactive agents with improved instruction-following ability. However, this can impact confidence calibration for reliable model output, which has not been researched in full. In this work, we examine various open-sourced LLMs, where we identify significant calibration degradation after instruction tuning. Seeking a practical solution, we look towards label smoothing, which has been shown as an effective method to regularize for overconfident predictions but has yet to be widely adopted in the supervised fine-tuning (SFT) of LLMs. We provide insight into why label smoothing can maintain calibration throughout the SFT process, but identify settings remain where the effectiveness of smoothing is severely diminished. We posit the cause to stem from the ability to become over-confident, which has a direct relationship with the hidden and vocabulary size of models, which we justify theoretically and experimentally. Finally, we address an outstanding issue regarding the memory footprint of the cross-entropy loss computation with label smoothing, designing a customized kernel to dramatically reduce memory consumption without sacrificing speed or performance in comparison to existing solutions.
\end{abstract}

\section{Introduction}

Tremendous progress has been made in building models that follow natural language instructions~\citep{it_human_feedback, prompting_zsg, it_scaling} through the use of LLMs pre-trained on large amounts of data as well as high-quality datasets that enable them to learn to interact in a human-like manner~\citep{promptsource, supernli, tulu}. However, such models have demonstrated a propensity for over-confidence in their predictions~\citep{contextual_calibration, calibration_qa, llms_confidence}, eliciting concerns over their use in more high-stakes decision-making scenarios. Such observations are not new with respect to neural networks, which have consistently been shown to suffer from over-confident predictions and over-estimate the likelihood of their correctness~\citep{Calibration_NN, inception, when_ls_work, ECE, revisiting_calibration}. To improve this, methods such as temperature scaling~\citep{Calibration_NN} and label smoothing~\citep{when_ls_work} have been proposed as solutions with varying effectiveness, spurring additional work in ensuring that predictions and confidence remain matching~\citep{focal_loss, focal_loss_calibration, penalizing_confidence, margin_based_label_smoothing}.

In this work, we focus on label smoothing (LS) for calibration in SFT settings. Motivated by previous works showing the effectiveness of LS for calibration in different settings, we first verify its effectiveness. We demonstrate that while it can be effective, problems begin to emerge in the case of large vocabulary LLMs (LV-LLMs). To explain this phenomenon, we attempt to establish a link between the predictive abilities of such LLMs and their size. We show that in such settings, an explicit link between the lower bound of the model entropy and the hidden size of the model causes models to fail to become overconfident~\citep{contextual_calibration}, negating the potential benefits and use of label smoothing. We further show how alternative techniques, such as temperature scaling and logit capping, explicitly act as a mechanism to steer models toward overconfidence, allowing the benefits of label smoothing to once again emerge.

Nevertheless, a problematic setting still remains. Growing vocabulary sizes cause large amounts of memory to be consumed to materialize the relevant logits and probabilities, making training difficult. While efficient methods~\citep{CCE_loss, liger, torchtune} have been proposed by implementing hardware-level optimization that significantly reduces this memory bottleneck, they often cannot support label smoothing. To address this, we identify specific optimizations that can be made in computing matrices in GPU memory, allowing for the support of label smoothing with minimal increases in memory or computational speed. Thus we introduce a new kernel that enables us to use label smoothing more efficiently as a whole.

To summarize our contributions, we: \textbf{1)} We point out that common SFT practices significantly degrade LLM model calibration. \textbf{2)} We demonstrate and justify why label smoothing is an appropriate approach to mitigating this concern. \textbf{3)} However, we further identify specific issues where label smoothing remains prohibitive, particularly with large vocabulary LLMs, and why existing methods fall short. \textbf{4)} We demonstrate that optimizations exist, which we incorporate in custom kernels which enable us to perform label smoothing with significant memory and throughput improvements without performance sacrifices.

\section{Related Work}

\paragraph{Model Calibration}\citep{statistical_calibration, murphy1972, rss1983} is a concept of matching the prediction probabilities yielded for different inputs to the expected accuracy on these inputs. In a $K$-way classification setting, let $\mathcal{X} \in \mathbb{R}^{D}$ and $\mathcal{Y} \in \{\gamma_k\}_{k=1}^K$ indicate the input and label space, respectively. Let $f$ be a classifier and $f\left(\hat{y}|\vx\right) = \hat{c}$ be the confidence of prediction, i.e., the maximum of probabilities among $K$ dimensions corresponding to its prediction $\hat{y}$. A model is \textit{perfectly-calibrated} when 
\begin{align}
    P\left(\hat{y} = y | \hat{c} = c\right) = c \ \ \forall c \in [0,1].
\end{align}
Qualitatively, model calibration can be derived as $\mathbb{E}\left[\left|{P}\left(\hat{y} = y | \hat{c} = c\right) - c\right|\right]$. One metric that has been widely used for measuring calibration is the expected calibration error (ECE)~\citep{ECE}, which is a weighted average of bin-wise mis-calibration. The ECE divides the confidence score of $N$ samples into $M$ uniform confidence bins $\{B_{m}\}_{m=1}^{M}$ and takes the mean of the gap between accuracy ($\mathrm{acc}$) and confidence ($\mathrm{conf}$) over the bins weighted by the number of samples in the bins. 
\begin{align}
   \text{ECE} = \sum_{m=1}^{M} \frac{\left|B_m\right|}{N}\left|\mathrm{acc}\left(B_{m}\right)-\mathrm{conf}\left(B_{m}\right)\right|. 
\end{align}

Additional metrics that have been proposed include the Root Mean Square Calibration Error (RMS-CE)~\citep{RMS_CE}, which places greater emphasis on large calibration deviations, and the Static and Adaptive Calibration Errors (SCE/ACE)~\citep{SCE_ACE}, which measure miscalibration over fixed and data-dependent binning schemes, respectively. These metrics offer complementary perspectives to standard calibration error measures, enabling a more comprehensive assessment of model confidence alignment. 
\paragraph{Label Smoothing} (LS) has been demonstrated to be a promising paradigm in settings to prevent models from becoming overconfident~\citep{inception, when_ls_work} or when noise exists in the provided labels~\citep{smoothing_noisy_labels, smoothing_noisy_labels2, LABO}. Consider a model parameterized by $\vtheta$ to model a conditional distribution $P(\cdot|\vx;\vtheta)$, where the final operation is a $\mathrm{softmax}$. Consider the model to apply a function $f(\cdot;\vtheta)$ on $\vx$ and $\hat{\vxigma}(\vx;\vtheta)\in[0, 1]^K$ to be the post-$\mathrm{softmax}$ output. Then
\begin{align}
    P(\gamma_i|\vx;\vtheta)=\hat{\vxigma}(\vx;\vtheta)_i=\frac{\exp(\vell(\vx)_i)}{\sum_{k=1}^K\exp(\vell(\vx)_k)},
\end{align}

where $\vell(\vx)\in\mathbb{R}^K$ is the pre-\texttt{softmax} output of the model, commonly referred to as the logits or log-probabilities. Models are usually trained by minimizing a cross-entropy (CE) loss on a dataset $\mathcal{D}=\{\vx_n, y_n\}_{n=1}^N$ sampled from an unknown distribution $p(\vx, y)$ in order to learn the true conditional distribution $p_{y|\vx}(y|\vx)$, computed as
\begin{equation}\label{eq:ce_loss}
    \begin{split}
        \mathcal{L}^{\text{CE}}_{\mathcal{D}}(\vtheta)
        &=-\frac{1}{N}\sum_{i=1}^N\sum_{k=1}^{K}\delta_{y_n}^{\gamma_k}\log P(\gamma_k|\vx;\vtheta) \\
        &\approx -{\mathbb{E}}_{p(\vx, y)}\left[\sum_{k=1}^Kp(\gamma_k|\vx)\log P(\gamma_k|\vx;\vtheta)\right] \\
        &= -{\mathbb{E}}_{p(\vx, y)}[\mathrm{KL}\left[\vxigma(\vx)\|\hat{\vxigma}(\vx;\vtheta)]\right] + c \\
        &= \mathcal{L}^{\text{CE}}_{p(\vx, y)}(\bm{\theta)},
    \end{split}
\end{equation}
where $\delta_i^j$ is the Kronecker delta with value $1$ only when $i=j$. Label smoothing mixes the original distribution with a discrete uniform distribution $\mathcal{U} = \left[1/K\right]^K\in \mathbb{R}^K$ using a smoothing rate $\beta\in \left[0, 1\right]$. The loss then becomes
\begin{equation}\label{eq:ls-loss}
    \begin{split}
        % \mathcal{L}^{\text{LS}}_{\mathcal{D}}(\vtheta;\beta)
        \mathcal{L}^{\text{LS}}_{\mathcal{D}}(\vtheta)
        &=-\frac{1}{N}\sum_{i=1}^N\left[\sum_{k=1}^{K}\left[(1-\beta)\delta_{y_n}^{\gamma_k}+\frac{\beta}{K}\right]\log P(\gamma_k|\vx;\vtheta)\right] \\
        &=(1-\beta)\mathcal{L}^{\text{CE}}_{\mathcal{D}}(\vtheta) + \frac{\beta}{K}\sum_{i=1}^N\mathrm{KL}[\vu\|\hat{\vxigma}(\vx_n;\vtheta)]+c \\
        &\approx-{\mathbb{E}}_{p(\vx, y)}\left[\mathrm{KL}\left[(1-\beta)\vxigma(\vx)+\beta\vu\|\hat{\vxigma}(\vx;\vtheta)\right]\right]+c \\
        &= \mathcal{L}^{\text{LS}}_{p(\vx, y)}(\vtheta).
    \end{split}
\end{equation}
Thus, label smoothing can be understood to a regularization term that encourages a uniform distribution over the output labels, hence preventing it from over-fitting to the training data and encouraging a model to be less confident on all samples by smoothing the true conditional being learned.

\section{Smoothing and Calibration in LLMs}\label{sec:background}

\paragraph{Preliminaries.} Define an auto-regressive LLM to be parameterized by parameters $\vtheta$. A model represents an embedding function $g(\cdot;\vtheta_e):\mathbb{R}^{N}\to\mathbb{R}^{D\times N}$ where $L$ is the length of a discrete input sequence $\vx$ and $D$ is the hidden size of the model that produces an embedding $\mE\in\mathbb{R}^{D\times N}$. This is followed by a classifier $\mC(\vtheta_c)\in\mathbb{R}^{D\times \left|\mV\right|}$ and a $\mathrm{softmax}$ operation to produce a probability distribution over $\mV$.

A common practice is to tune $\vtheta = [\vtheta_e,\vtheta_c]$ on a dataset through supervised fine-tuning (SFT), or instruction tuning~\citep{sft_models_zero_shot_learners, it_human_feedback, it_scaling, alpaca}. During SFT of an LLM, an input sequence $\vx$ of length $L$ consists of a sequence of discrete vocabulary tokens $\vx_i\in\mV \ \forall i \in[N]$. The first $m$ tokens in $\vx$ consist of the instruction, while the rest is considered the target output. The goal of SFT is to minimize a CE objective over the output portion of the sequence $\vx_{m+1:N}$, computed as a loss over the individual elements of the sequence

\begin{align}
   \mathcal{L}_{\vx}(\vtheta)=\sum_{j, v}  \delta_{\vx_j}^{v}\log\left( \mC(\vtheta_c)^\top g(v|\vx_{1:j-1};\vtheta_e)\right). 
\end{align}

In essence, the learning problem is a $\left|\mV\right|$-class classification problem for each element in the target portion of the input sequence, where the prediction for any specific position is influenced by all previous elements in the sequence.

\begin{figure}[t!]
    \centering
    \includegraphics[width=\linewidth]{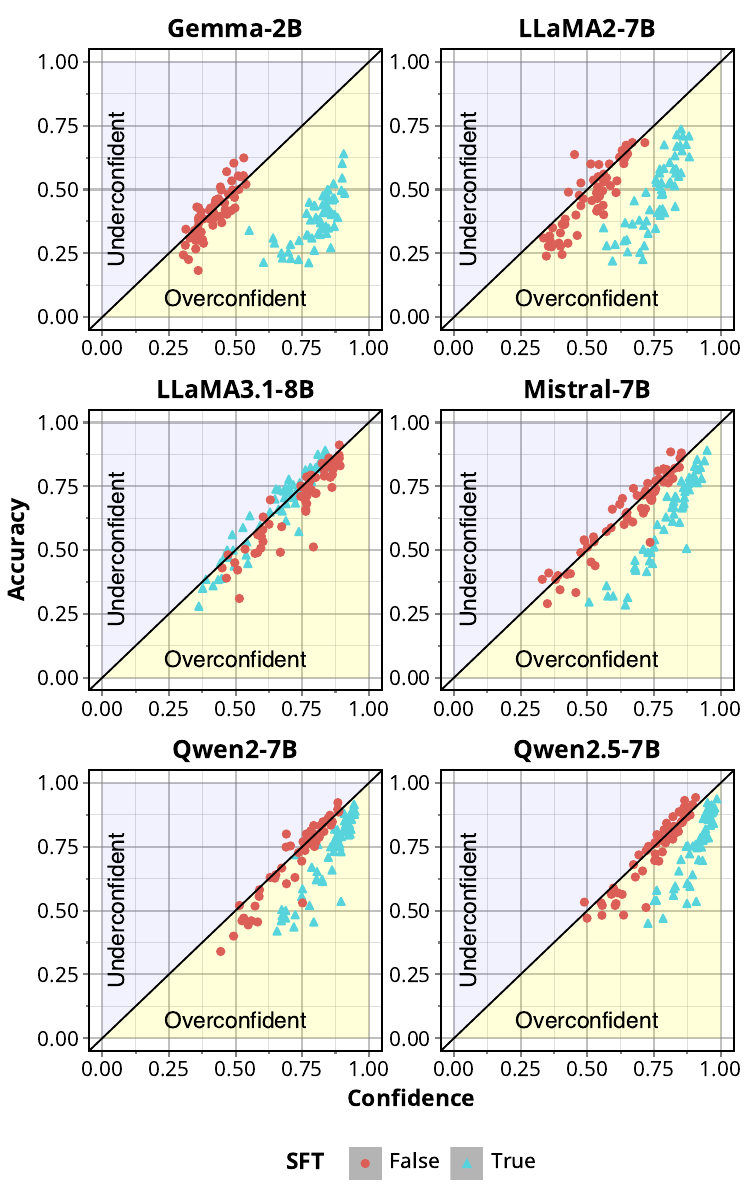}
    \vspace{-0.5cm}
    \caption{Reliability diagrams of open-sourced pre-trained models with (red) and without instruction-tuning (blue) on the MMLU dataset~\citep{mmlu}. The horizontal axis represents the model’s confidence in each answer choice for each question, while the vertical axis shows the accuracy on each question. The solid diagonal indicates perfect calibration, separating areas where predictions are deemed over-confident or under-confident. Instruction-tuning visibly leads to over-confidence, regardless of the instruction-tuning dataset (which differs between models). 
    }
    \label{fig:poor_calibration}
\end{figure}

This procedure transforms the original parameter set $\vtheta$ into a new set, denoted $\vtheta_{\text{SFT}}$, which often enables the model to follow human-provided instructions with remarkable accuracy and fluency. However, this fine-tuning process can also lead to a deterioration in the model’s calibration, as illustrated in \cref{fig:poor_calibration}, potentially reducing its reliability in estimating uncertainty or confidence.

\paragraph{Why does Instruction Tuning Lead to Mis-calibration?} Our results raise a question: \textit{Why does instruction tuning degrade calibration?} To better understand this, SFT can be viewed through the lens of out-of-distribution (OOD) generalization and calibration error. In particular, we assume that the SFT data consists of an in-distribution (ID) dataset whereas the downstream dataset on which generalization and calibration are tested constitutes an OOD dataset.

Suppose we have an ideal parameter set $\vtheta^*$ that minimizes calibration error on an unknown dataset $\mathcal{D}$, defined as $\mathbb{E}_{(\vx,y)\sim\mathcal{D}}[\|f(\vx;\vtheta) - c(\vx)\|^2_2]$ where $c(\vx) = \mathbb{E}_{y\sim f(\vx;\vtheta)}[y]$ is the expected label given a prediction $f(\vx;\vtheta)$. In other words, $f(\cdot;\vtheta)$ always produces the calibrated prediction for the label given an input $\vx$ such that the output confidence matches the expected label over the subset of samples with the same confidence value. Thus the goal of SFT is to learn a set of parameters $\vtheta^*$ that outputs reliable prediction probability on samples from both an unseen OOD domain, which is defined by a distribution $p_{\text{OOD}}(\vx, y)$, as well as the ID distribution $p_{\text{ID}}(\vx, y)$. Thus the goal is to minimize
\begin{align}
\mathcal{L}_{p}(\vtheta)=\mathbb{E}_{p}[\|f(\vx;\vtheta) - f(\vx;\vtheta^*)\|_2^2], 
\end{align}
for $p\in\{p_{\text{OOD}}(\vx, y),p_{\text{ID}}(\vx, y)\}$. 

\begin{figure*}[th!]
    \centering
    \includegraphics[width=\linewidth]{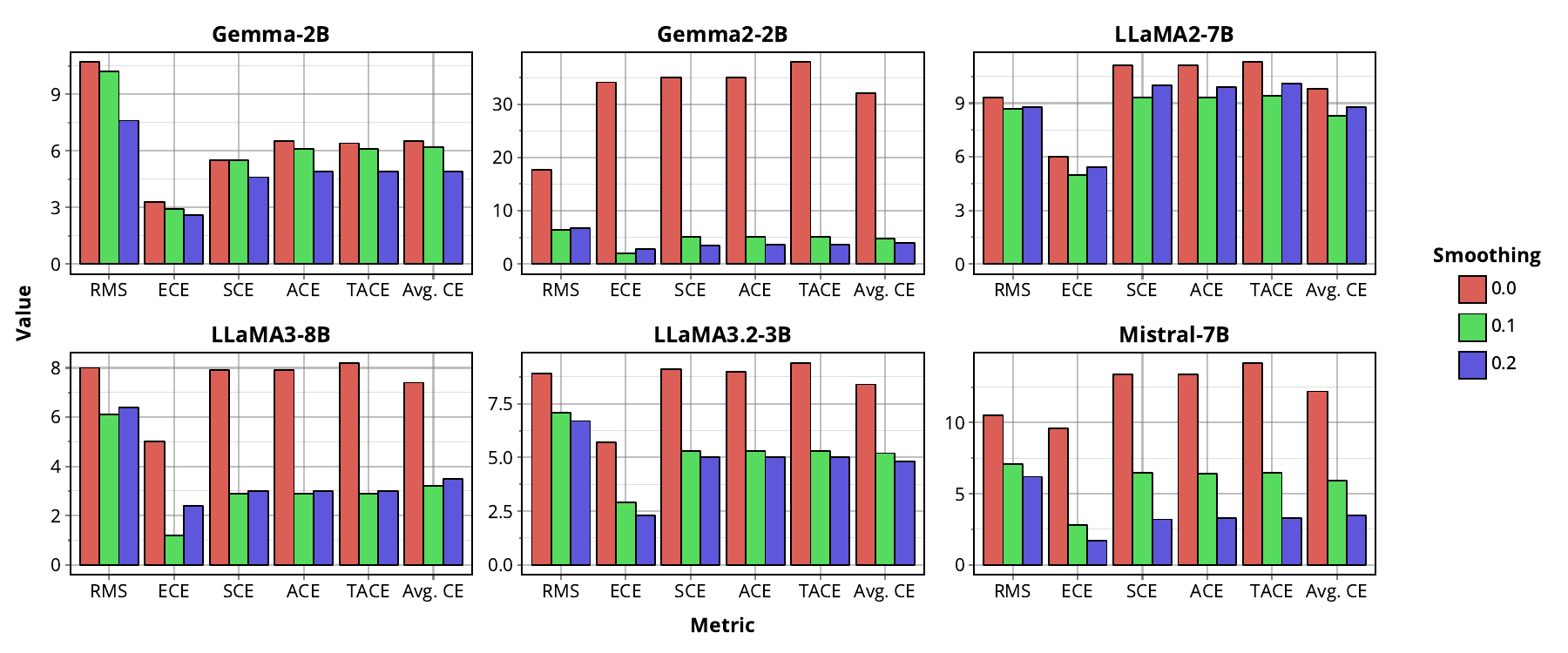}
    \vspace{-0.5cm}
    \caption{Effects of instruction-tuning on calibration, presented under a number of different calibration error metrics (where lower is better). Values can range from 0 to 100. Models are all fine-tuned on a \texttt{Tulu3}~\citep{tulu} SFT dataset and evaluated on MMLU. We can observe that across all models, which have various structural differences, the use of label smoothing is capable of reducing calibration error while having negligible effects on downstream performance accuracy on the task.}
    \label{fig:sft-miscalibration}
\end{figure*}
\citet{calibrated_finetuning} demonstrate that under such a setting, simultaneously maintaining accuracy and calibration of the final classifier (in the case of an auto-regressive LLM, this is the language modeling head) has a direct relationship to the diversity of the feature embeddings as follows:
\begin{lemma}[\citealt{calibrated_finetuning}]\label{theorem:calibrated_finetuning} %这还要想怎么adapt到multi-class的情况
    Let $f(\cdot;\vtheta):\mathcal{X}\rightarrow[0,1]^K$ be a real-valued function of the form $f(\vx;\vtheta)=\sum_{i=1}^{d}f_{i}(\vx[i];\vtheta)$ where $f_{i}(\cdot;\vtheta)$ is an arbitrary one-dimensional function, and $f$ is in a hypothesis class $\mathcal{F}$ that has pseudo dimension $\mathcal{P}_{\text{dim}}(\mathcal{F})=d_{f}$. Let $\mathcal{D}_\text{ID} = \left\{\left(\vx^{\left(n\right)}_{\text{ID}}, y^{\left(n\right)}_{\text{ID}}\right)\right\}_{n=1}^N$ be a dataset sampled from the ID distribution. If $(\vx[1],\dots,\vx[d])$ have matching marginals for ID and OOD, and $(\vx[i],\vx[j])$ is a bi-variate Gaussian for every $i,j \in [d]$, then for any $\delta \in (0,1)$ and for all $f$, the following bounds hold with probability at least $1-\delta$:
    \begin{equation}\label{eq:ood_class_error}
        \begin{split}
        \mathcal{L}_{p_{\text{OOD}}}(\vtheta) &\le \mathcal{L}_{{\mathcal{D}}_{\text{ID}}}(\vtheta) + {\frac{d}{\sigma_{\text{min}}\left(\tilde{\Sigma}_{p_{\text{ID}}\left(\vx\right)}\right)}} + \Delta\\
        &\quad+ \mathcal{O} \left(\sqrt{\log{\left(\frac{N}{d_{h}}\right)}^{d_{h}}\left(\frac{1}{N\delta}\right)}\right),
        \end{split}
    \end{equation}
    \begin{equation}\label{eq:ood_cal_error}
        \begin{split}
        &\mathbb{E}_{p_{\text{OOD}}(\vx, y)}\left[\left\|f\left(\vx;\vtheta\right)-y\right\|_2^{2}\right] + \mathbb{E}_{p_{\text{OOD}}(\vx, y)}\left[\left\|c\left(\vx\right)\right\|_2^{2}\right] - 1 \\
        \le &\mathcal{L}_{{\mathcal{D}}_{\text{ID}}}(\vtheta) + {\frac{d}{\sigma_{\text{min}}\left(\tilde{\Sigma}_{p_{\text{ID}}(\vx)}\right)}} + \Delta \\
        &+ \mathcal{O} \left(\sqrt{\log{\left(\frac{N}{d_{h}}\right)^{d_{h}}}\left(\frac{1}{N\delta}\right)}\right),
        \end{split}   
    \end{equation}
    where
    \begin{equation}
        \begin{split}
            \theta^*&=\underset{\vtheta\in{\Theta}}{\mathrm{argmin}}\phantom{0}\mathcal{L}_{p_{\text{OOD}}(\vx, y)}(\vtheta)+\mathcal{L}_{p_{\text{ID}}(\vx, y)}(\vtheta), \\
            \Delta&=\mathcal{L}_{p_{\text{OOD}}(\vx, y)}(\vtheta^*)+\mathcal{L}_{p_{\text{ID}}(\vx, y)}(\vtheta^*),
        \end{split}   
    \end{equation}
    and $\tilde{\Sigma}_{p_{\text{ID}}(\vx)}=\mathbb{E}_{p_{\text{ID}}(\vx)}[\tilde{\vx}\tilde{\vx}^{\top}]$ is a covariance matrix with a strictly positive minimum singular value of $d$-dimensional normalized input $\tilde{\vx}=(\tilde{\vx}[1],...,\tilde{\vx}[d])$, where $\tilde{\vx}[i]{=}(\vx[i] - \mathbb{E}[\vx[i]])\sqrt{\text{Var}(\vx[i])}$ and $\sigma_{\text{min}}(M)$ is the smallest singular value of a matrix $M \in \mathbb{R}^{d_{1}\times d_{2}}$. 
\end{lemma}

The dependence of the bound on the minimal singular value of the covariance matrix indicates that as the set of learnt feature embeddings (the embedding of the context in this scenario) becomes less mutually dependent, both calibration error and classification error can be minimized. However, prior works have shown that fine-tuning can significantly reduce the diversity of such features~\citep{finetuning_cripple, finetuning_distort, platonic_representations}, justifying why standard SFT can significantly degrade calibration (\cref{fig:sft-miscalibration}).

\paragraph{The Effects of Label Smoothing.} To understand the effects of label smoothing from a model calibration perspective, we adopt an optimization-based viewpoint to analyze the implicit constraints imposed by the regularization~\citep{nonlinear_programming}. This perspective reveals how label smoothing modifies the solution space of the model’s predictions, encouraging not only smoother output distributions but also a reduction in the model’s confidence on any single class.  First, define
\begin{definition}\label{def:logit_distance}
    The \textbf{logit distance} vector for $\vx$, $\bm{d}(\vx)$, is
    \begin{equation}\label{eq:logit_distance}
        \bm{d}(\vx) = \left[\max_{1\leq i\leq K}\vell(\vx)_i - \vell(\vx)_k\right]_{k=1}^K\in\mathbb{R}^K.
    \end{equation}
\end{definition}
One way of ensuring that a model does not over-estimate a specific class is to enforce this as a hard constraint, which results in equal logits among all classes and a $\mathrm{softmax}$ output of $\bm{o}=f(\vx;\vtheta)=[1/K]^K$. As such, it is often preferable to enforce this as a soft-penalty function $\mathcal{P}: \mathbb{R}^K\to\mathbb{R}$ into the objective function minimized during training. Recalling \cref{eq:ls-loss}, we can relate this soft-penalty to the additional KL-divergence introduced by the label smoothing objective.

\begin{proposition}\label{prop:consraint}
    A linear penalty (or a Lagrangian term) for the hard constraint $\bm{d}(\vx) = \bm{0}$ is bounded from above and below by $\mathrm{KL}\left(\vu\|\hat{\vxigma}\left(\vx;\vtheta\right)\right)$, up to additive constants
    \begin{equation}
        \mathrm{KL}[\vu\|\hat{\vxigma}\left(\vx;\vtheta\right)]-\log K \leq \sum_{i=1}^K\frac{\bm{d}\left(\vx\right)_i}{K}\leq \mathrm{KL}\left[\vu\|\hat{\vxigma}\left(\vx;\vtheta\right)\right].
    \end{equation}
\end{proposition}
The proof (in \cref{sec:constraint_proof}) indicates that label smoothing approximately minimizes, for a linear penalty, the constraint $\bm{d}(\vx)=\bm{0}$, encouraging equality among the logits for each class to ensure that over-confidence is penalized. More importantly, however, it is noted that

\begin{proposition}\label{prop:ls-map}
    Define a likelihood model $p\left(y | \vx; \vtheta\right) = \mathrm{Cat}\left( \mathrm{softmax}\left(f\left(\vx;\vtheta\right)\right) \right)$, a categorical distribution with parameters $\vz=\mathrm{softmax}\left(f\left(\vx;\vtheta\right)\right)  \in \Delta(\Theta)$ where $\Delta\left(\Theta\right)$ denotes a probability simplex over the parameter space $\Theta$.
    The label smoothing objective is equivalent to Maximum A Posteriori (MAP) estimation on the $\mathrm{softmax}$ probability vector under the independence assumption $p\left(\vz|\vx\right) = p\left(\vz\right)$.
\end{proposition}
A proof is provided in \cref{sec:proof-ls-map}. This MAP formulation above relies on the provided label $y$ for each sample $\vx$, without exploiting the potential similarities among different samples in the empirical training dataset for more accurate estimation. \citet{analysisofmap} showed that using MAP estimation can lead to greater separability and diversity of individual samples, under the assumption that the sample is sampled from a normal distribution. \citet{logitsnormal} further prove this to be the case with Transformer-based language models. In conjunction with prior claims from \citet{calibrated_finetuning}, this proposition indicates that label smoothing can in fact learn more diverse input features, further explaining the improvement in calibration.

\section{Label Smoothing for Large Vocabularies}

\begin{figure}[th!]
    \centering
    \includegraphics[width=0.75\linewidth]{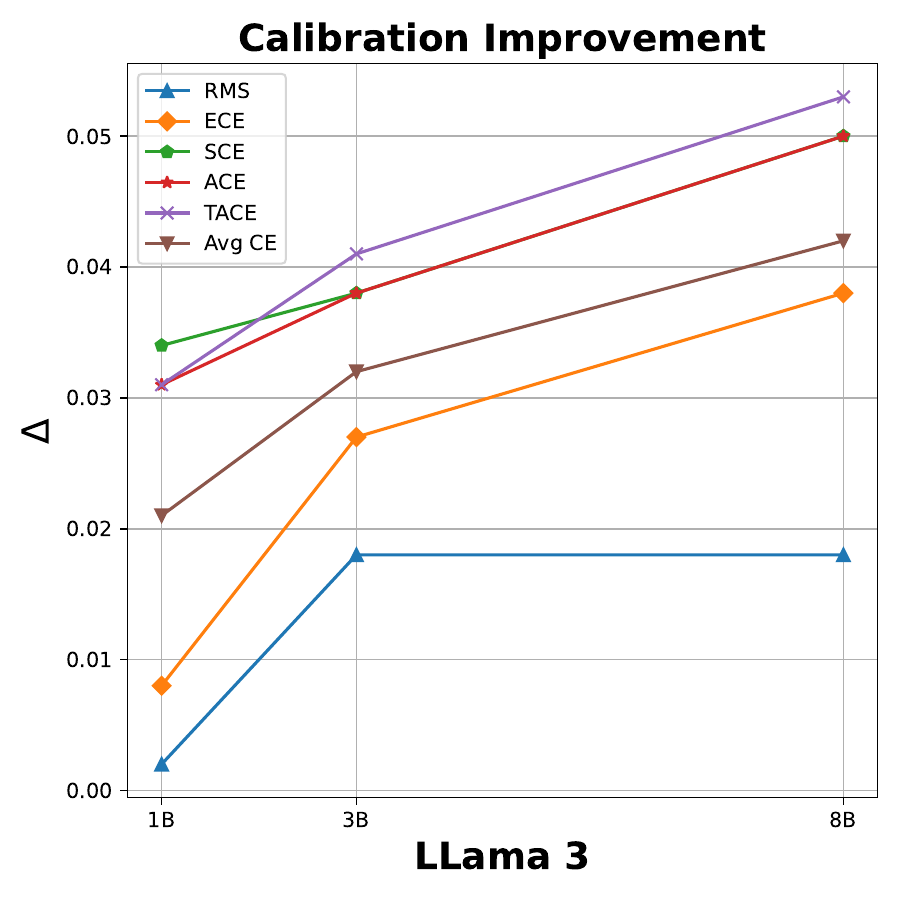}
    \vspace{-0.5cm}
    \caption{Calibration of different \texttt{LLaMA3} models fine-tuned on the same SFT dataset. As the size of the model decreases, the calibration of the model sees less improvement from the use of LS.}
    \label{fig:calibration_llama_tulu3mixture}
\end{figure}

We conduct SFT training with and without LS on a Tulu3 dataset~\citep{tulu} for different pre-trained language model families, including Llama~\citep{llama3}, Gemma~\citep{gemma2} and Mistral~\citep{mistral}. 
While we can note the usefulness of label smoothing for model calibration shown in \cref{fig:sft-miscalibration}, it becomes clear that its effectiveness is much less visible in some cases. Take for instance three \texttt{LLaMA3} models of sizes 1\texttt{B}, 3\texttt{B} and 8\texttt{B}, which we fine-tune on the same instruction dataset (\cref{fig:calibration_llama_tulu3mixture}). While label smoothing shows an improvement in calibration for the 8\texttt{B}-sized model, this diminishes significantly to the 3\texttt{B} model and 1\texttt{B} model, both in the baseline model (where no SFT procedure has been performed) as well as SFT with multiple different datasets. We therefore seek to investigate the underlying causes of our empirical findings, aiming to shed light on a previously unexplained phenomenon.

We study the concentration behavior of the LM head by analyzing its relationship between entropy and model size, which provides an explanation of diminishing returns of label smoothing. We start by providing the following lemma, which establishes a bound on the LLM logits prior to the language modeling head.

% \subsubsection{The Relation Between Size and Entropy}

\begin{figure}[t!]
    \centering
    \includegraphics[width=0.9\linewidth]{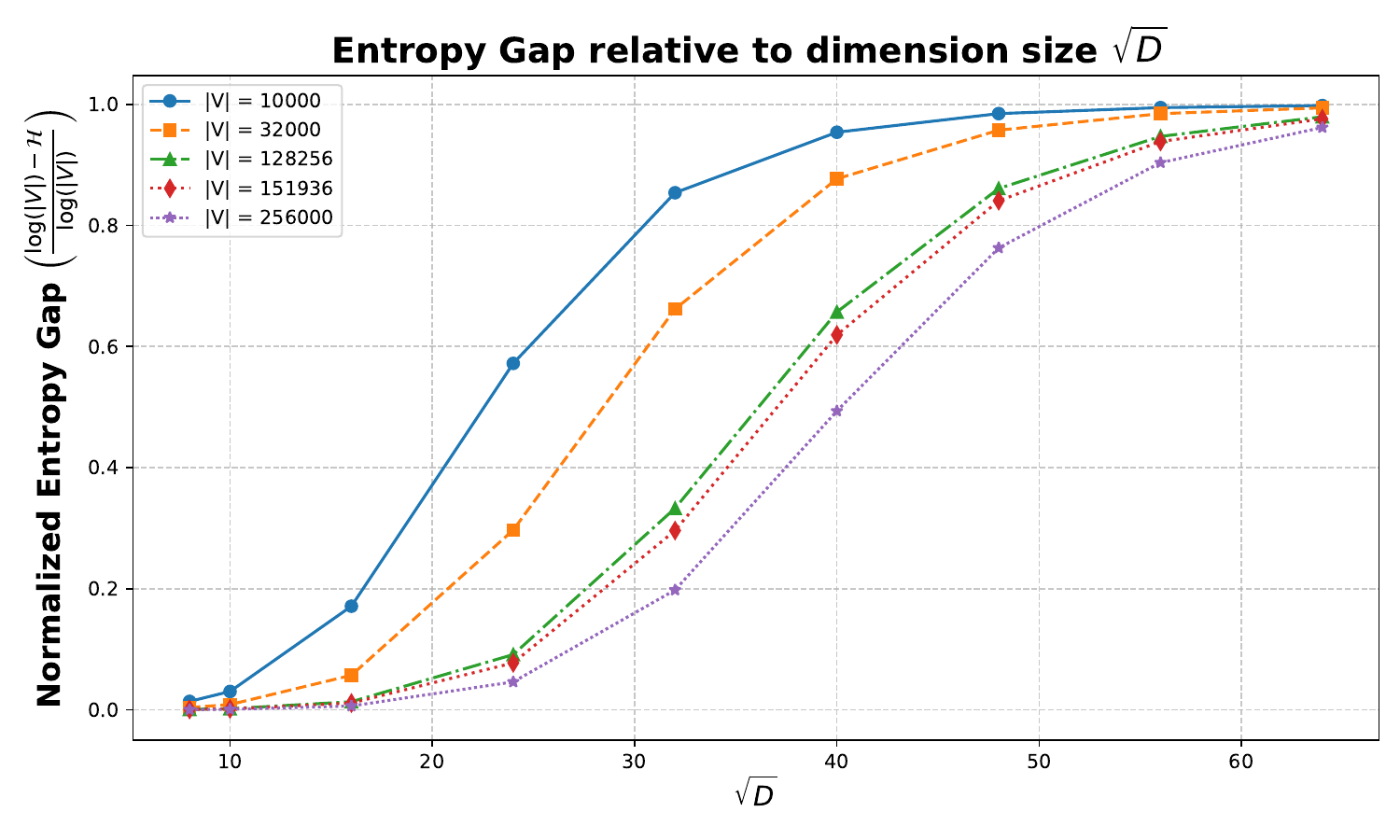}
    \vspace{-0.5cm}
    \caption{Relative entropy bound for different LLM vocabulary sizes with varying hidden sizes ($D$). Our visualization shows the normalized entropy gap for varying hidden sizes of the LM head. This gap is calculated by taking the difference between the entropy upper and lower bounds and dividing by the upper bound ($\log\lrbracknorm{\mV}$). A lower ratio indicates the model is restricted to producing concentrated predictions. }
    \label{fig:lower_bound}
\end{figure}

\begin{lemma}\label{lemma:norm}
    Let \(\mC \in \mathbb{R}^{D \times \lrbracknorm{\mV}}\) be a matrix with \(\lrbrackvec{\mC}_2 = \sigma_C\), and let \(\vh \in \mathbb{R}^D\) be a vector such that each entry of \(\vh\) satisfies \(\left|h_i\right| \leq \sigma_h\). The upper bound for \(\left\|\mC^\top \vh\right\|_2\) is:
\begin{align}
    \lrbrackvec{\mC^\top\vh}_2 \leq \sigma_C \cdot \sigma_h \cdot \sqrt{D}.
\end{align}
\end{lemma}
Thus, the norm of the final logit vector is also bounded and depends on the dimension dimension, which enables an entropy-based analysis of the LM head's predictions as a function of both the vocabulary size $\lrbracknorm{\mV}$ and the embedding dimension $D$.

\begin{theorem}\label{thm:entropy_bound}
    (LM head Entropy Lower Bound). Let $\rho = \sigma_C\sigma_h$, $\vu=\mC^\top\vh$ and $\gamma = \exp\lrbrackround{-\rho\sqrt{\frac{D\lrbracknorm{\mV}}{\lrbracknorm{\mV}-1}}}$, then the entropy $\mathcal{H}_{\vu}$ of prediction of the LM head holds that:
    {\small\begin{align}
        \mathcal{H}_{\vu} \geq \log\lrbrackround{1 + \lrbrackround{\lrbracknorm{\mV}-1}\gamma} + \frac{\rho\cdot\gamma\sqrt{D\lrbracknorm{\mV}\lrbrackround{\lrbracknorm{\mV}-1}}}{1 + \lrbrackround{\lrbracknorm{\mV}-1}\gamma}.
    \end{align}}
\end{theorem}
Intuitively, the above lemma and theorem (proofs in~\cref{subsec:norm_proof}) indicates that for large $\lrbracknorm{\mV}$, the minimum entropy is lower bounded by $\Omega\lrbrackround{\lrbracknorm{\mV}e^{-\rho\sqrt{D}}}$, thus increasing linearly with the vocabulary size $\lrbracknorm{\mV}$ and decreasing exponentially with $\sqrt{D}$. We illustrate how this changes in \cref{fig:lower_bound}, where the normalized entropy gap for different $\lrbracknorm{\mV}$ is shown. Given the same $\lrbracknorm{\mV}$, the concentration behavior of the LM head is primarily influenced by the size of the hidden dimension. As the hidden size increases, the model is increasingly capable of attaining a lower entropy, while the bound is smaller for larger $\lrbracknorm{\mV}$ at the same $D$, highlighting why large vocabulary LLMs at smaller sizes are less prone to overconfidence during tuning.
\begin{remark}
Models with smaller $D$ with large vocabulary size $\left|\mV\right|$ suffer from a lack of concentration ability due to their limited hidden size. By consequence, label smoothing cannot help with calibration, as it serves to only penalize overconfidence while having no specific benefits for under-confidence due to mixing with the maximum entropy (uniform) distribution.
\end{remark}

However, we note that from this analysis, the entropy bound is influenced by $\lrbracknorm{\mV}$ and $D$ through the distribution of $\vh$. However, as $\vh$ are the logits of the model, these can be manipulated before the $\mathrm{softmax}$ directly. Thus it follows that under-confidence in models can be attained through manipulation of the $\lrbrackvec{\mC^\top\vh}$:
\begin{remark}
    Fix $\lrbracknorm{\mV}$ and $D$. Using a temperature $\tau < 1$ will modify $\rho$, leading to a decreased lower bound.
\end{remark}

We note that various methods can serve to manipulate this bound. The above remark notes the effects of temperature scaling, which directly uses a constant temperature $T$ to scale the logits before the $\mathrm{softmax}$. This enables the manipulation of $\sigma_h$ in \cref{lemma:norm} without changing any additional values, thereby increasing or decreasing the entropy bound based on the choice of $T$. This explains how temperature scaling can enable better model calibration, especially through the choice of a $T > 1$ used to divide the logits, thereby decreasing $\sigma_h$ which has the downstream effect of increasing the minimum entropy bound of the model. This can (for smaller label class sizes) significantly improve calibration, as the initial model has the potential for over-confidence which temperature can serve to mitigate. Similarly, the use of logit softcapping, as in the \texttt{Gemma2} family of models, applies a similar change in $\sigma_h$, reducing the lower bound and thereby enabling models of smaller size to become over-confident.

\begin{figure}[th!]
    \centering
    \includegraphics[width=0.9\linewidth]{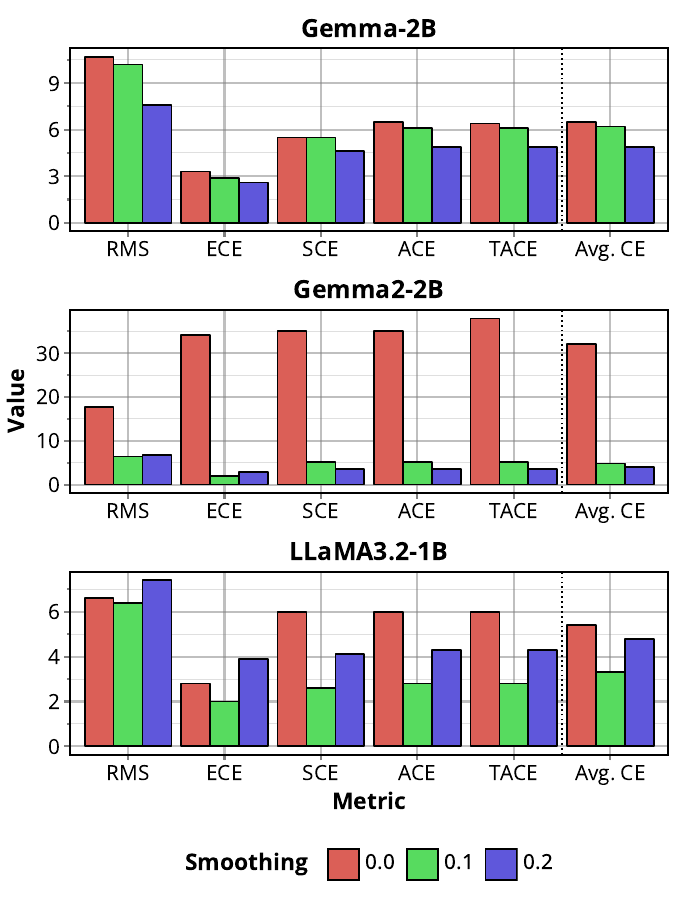}
    \vspace{-0.5cm}
    \caption{Effect of label smoothing on large vocabulary models with a smaller hidden size (2048). \texttt{Gemma-2B} observes a smaller change compared to \texttt{LLaMA3.2-1B}, due to having the largest vocabulary size. However, \texttt{Gemma2-2B} observes a large change in part thanks to the softcapping of logits.}
    \label{fig:calibration-small}
\end{figure}

\section{Efficient Smoothed Cross-Entropy}

Despite our analysis on the effectiveness of label smoothing for calibrating LLMs, areas of concern still exist regarding practical applicability. In this section, we introduce such limitations and how the label smoothing setting is distinct from existing solutions that attempt to mitigate them.

When the vocabulary $\mV$ is large, the final cross-entropy begins to consume a significant amount of memory, as a $N\times\lrbracknorm{\mV}$ matrix must be materialized to compute the loss. While optimizations have been implemented to manage other model components~\citep{fla, fla2}, this logit matrix begins to constitute the vast majority of this footprint when the vocabulary is large~\citep{llama3, gemma2, qwen2.5}. In this section, we introduce an efficient smoothed cross-entropy computation, with its core implementation and results that demonstrate its effectiveness.

\subsection{Limitations of Existing Solutions} 
Efficient implementation of cross-entropy computation~\citep{CCE_loss, liger, torchtune} often makes use of the fact that for a non-smoothed loss, computing the entire logit matrix is unnecessary. Instead, only the relevant row from the classifier head $\mC$ is needed, as this is the only raw logit value that is relevant to the loss. Such approaches therefore store into an output only this single logit, reducing overhead by a factor of nearly $\frac{1}{\lrbracknorm{\mV}}$. However, this is not feasible for smoothed losses. Reusing notation from \cref{sec:background}, consider an LS loss on a sequence
\begin{equation}\label{eq:ls-breakdown}
    \begin{split}
    &\mathcal{L}^{\text{LS}}_{\vx} = \sum_{i=1}^N\mathcal{L}^{\text{LS}}_{\vx_i} = \sum_{i=1}^N 
    \bigg[
    \underbrace{\left(1-\beta\right)\mC_{\vx_{i}}^\top \mE_i}_{\mone\text{Target Loss}} \\
    &+\underbrace{\frac{\beta}{\left|\mV\right|}\sum_{v\in\mV}\mC_{v}^\top \mE_i}_{\mtwo\text{Smoothing Loss}}-\underbrace{\log\sum_{v\in\mV}\exp\left(\mC_{v}^\top \mE_i\right)}_{\mthree\mLSE}
    \bigg],
    \end{split}
\end{equation}
$\mtwo$ in the above equation indicates that it must be the case that all logits need to be explicitly added to the loss, leaving existing solutions unfeasible for this specific setting. However, we demonstrate that we can indeed compute these in a manner that limits memory consumption usage, without influencing the throughput and applicability of the proposed solution to other scenarios such as no label smoothing.
\subsubsection{Forward Pass} To naively compute a CE loss as in~\cref{eq:ls-breakdown}, one could individually compute each loss component. However, we notice that many of these computations can be reused, such as the computation of $\mC^\top _v\mE_i$ in both $\mtwo$ and $\mthree$. As such, for more efficient computation, we can compute all components in parallel in on-chip shared memory (SRAM) while making the best use of the GPU cache structure.
\subsection{Implementation}

\begin{algorithm}[ht!]
    \begin{tabularx}{\textwidth}{lX}
      \textbf{Inputs:} & $\mE \in \mathbb{R}^{{D}\times{N}}, \mC \in \mathbb{R}^{{D}\times{\left|\mV\right|}},\vx \in \mathbb{R}^{N}$.\\
      & Block sizes $N_B$, $V_B$, and $D_B$.\\
      & Smoothing $\beta\in[0,1]$.\\
      \textbf{Outputs:} &  $\mLSE = \log\sum_v\exp(\mC^\top_v\mE) \in \mathbb{R}^N$.\\
      & $\vo=
      (1-\beta)\left(\mC^\top\mE_i\right)_{\vx}+\frac{\beta}{\left|\mV\right|}\sum_{v}\mC_{v}^\top \mE \in \mathbb{R}^N.$
    \end{tabularx}
    \vspace{0.25em}
    \hrule
    \vspace{0.25em}
    \begin{algorithmic}
    \State $\mLSE,\vo = -\mathbf{\infty}_N,\mathbf{0}_N$ % \Comment{$-\infty$ and zero vectors of size $N$ in main GPU memory}
    \For{all pairs of blocks $\mE_n$, $\mC_v$, ${\vx}_n$} % \Comment{Divide $\mE,\mC, \vx$ into blocks of size ${D}\times{N_B},{D}\times{V_B}, N_B$}
        \State $\mA_{nv} = \mathbf{0}_{{V_B}\times{N_B}}$ % \Comment{Zero matrix of size ${V_B}\times{N_B}$ in on-chip SRAM}
        \For{blocks $\mE_{n,d}$, $\mC_{v,d}$} % \Comment{Divide $\mE_n$ and $\mC_v$ into blocks of ${D_B}\times{N_B}$ and ${D_B}\times{V_B}$}
            \State $\mA_{nv} \mathrel{+}= \mC_{v,d}^\top\mE_{n,d}$ % \Comment{Blockwise matrix multiplication}
            % \If{$\mC_{\vx_n,d}$ overlaps with $\mC_{v,d}$}
            %     \State $\vo_n\mathrel{+}= (1-\beta)\mE_{n,d}\cdot\mC_{\vx_n,d}$
            % \EndIf
            \State $\mM = \left(\mC_{\vx_n,d}==\mC_{v,d}\right)$ % \Comment{$0/1$-mask based on pointer address comparison}
            \If{$\mM\neq\mathbf{0}_{N_B\times V_B}$}
            \State $\vo_n\mathrel{+}= (1-\beta)\cdot\sum\left[\left(\mC_{m,d}^\top\mE_{n,d}\right)^\top\odot \mM\right]$ % \Comment{Regular loss accumulation}
            \EndIf
        \EndFor
        \State $\mLSE_{nv} = \log \sum \exp\left(\mA_{nv}^\top\right)$ % \Comment{Numerically stable implementation with max}
        \State $\mLSE_{n} = \log\left(\exp\left(\mLSE_{n}\right) + \exp\left(\mLSE_{nv}\right)\right)$ % \Comment{Locking thread-safe log-add-exp}
        \If{$\beta\neq0$}
            \State $\vo_n \mathrel{+}= \frac{\beta}{\lrbracknorm{\mV}}\sum \mA_{nv}^\top$ % \Comment{Smoothed loss accumulation}
        \EndIf
    \EndFor
    \end{algorithmic}
    \caption{Memory-efficient forward pass}
    \label{alg:combined}
\end{algorithm}

\begin{table*}[t!]
    \centering
    \caption{Results of different models w/ or w/o LS on different datasets. All models are performed with a 5-shot evaluation. We report the reliability of models with expected calibration error (ECE) and root mean square calibration error (RMS).}
    \label{tab:main_performance}
    \resizebox{0.9\linewidth}{!}{
        \begin{tabular}{llccccccccc}
        \toprule
        \multirow{2}{*}{\text{\textbf{SFT Dataset}}} & 
        \multirow{2}{*}{\text{\textbf{Model}}} & 
            \multicolumn{3}{c}{\textsc{\textbf{MMLU}}} & 
            \multicolumn{3}{c}{\textsc{\textbf{HellaSwag}}} & 
            \multicolumn{3}{c}{\textsc{\textbf{Arc-easy}}} \\
        \cmidrule(lr){3-5} \cmidrule(lr){6-8} \cmidrule(lr){9-11} 
        & & 
        \multicolumn{1}{l}{Acc. $\uparrow$} & \multicolumn{1}{l}{ECE $\downarrow$} & \multicolumn{1}{l}{RMS $\downarrow$} & 
        \multicolumn{1}{l}{Acc. $\uparrow$} & \multicolumn{1}{l}{ECE $\downarrow$} & \multicolumn{1}{l}{RMS $\downarrow$} & 
        \multicolumn{1}{l}{Acc. $\uparrow$} & \multicolumn{1}{l}{ECE  $\downarrow$} & \multicolumn{1}{l}{RMS $\downarrow$} \\
        \midrule
        \multirow{6}{*}{\texttt{Alpaca}} & \texttt{Mistral-7B} + SFT $(\beta=0)$ 
        & 0.579 & 0.134 & 0.120 
        & 0.302& 0.127& 0.160
        & 0.803& 0.099& 0.154\\
        & \texttt{Mistral-7B} + SFT $(\beta=0.1)$ 
        & 0.590 & \textbf{0.094} & \textbf{0.104}
        & 0.304& \textbf{0.087}& \textbf{0.124}
        & 0.806& \textbf{0.071}& \textbf{0.131}\\
        \cmidrule{2-11}
        & \texttt{LLaMA3-8B} + SFT $(\beta=0)$ 
        & 0.638 & 0.113 & 0.113
        & 0.375& 0.162& 0.085
        & 0.863& 0.070& 0.127\\
        & \texttt{LLaMA3-8B} + SFT $(\beta=0.1)$ 
        & 0.636 & \textbf{0.073} & \textbf{0.094} 
        & 0.374& \textbf{0.087}& \textbf{0.037}
        & 0.864& \textbf{0.037}& \textbf{0.098}\\
        \cmidrule{2-11}
        & \texttt{Gemma2-2B} + SFT $(\beta=0)$ 
        & 0.528 &0.343 & 0.180  
        & 0.302& 0.127& 0.160
        & 0.773& 0.131& 0.174\\
        & \texttt{Gemma2-2B} + SFT $(\beta=0.1)$ 
        & 0.532 & \textbf{0.125} & \textbf{0.121}
        & 0.304& \textbf{0.087}& \textbf{0.124}
        & 0.764& \textbf{0.069}& \textbf{0.127}\\
        \midrule
        \multirow{6}{*}{\texttt{Tulu3Mixture}} & \texttt{Mistral-7B} + SFT $(\beta=0)$ 
        & 0.600 & 0.096 & 0.105 
        & 0.369& 0.044& 0.085 
        & 0.843& 0.078& 0.135\\
        & \texttt{Mistral-7B} + SFT $(\beta=0.1)$ 
        & 0.603 & \textbf{0.028} & \textbf{0.071}
        & 0.375& \textbf{0.021} &\textbf{ 0.067}
        & 0.840& \textbf{0.030}& \textbf{0.094}\\
        \cmidrule{2-11}
        & \texttt{LLaMA3-8B} + SFT $(\beta=0)$ 
        & 0.651 & 0.050 & 0.080
        & 0.361& 0.049& 0.091 
        & 0.857& 0.058& 0.114\\
        & \texttt{LLaMA3-8B} + SFT $(\beta=0.1)$ 
        & 0.646 & \textbf{0.012} & \textbf{0.061} 
        & 0.356& \textbf{0.025}& \textbf{0.064}
        & 0.858& \textbf{0.035}& \textbf{0.097}\\
        \cmidrule{2-11}
        & \texttt{Gemma2-2B} + SFT $(\beta=0)$ 
        & 0.533 &0.341 & 0.177  
        & 0.273& 0.082& 0.128
        & 0.758& 0.086& 0.142\\
        & \texttt{Gemma2-2B} + SFT $(\beta=0.1)$ 
        & 0.531 & \textbf{0.020} & \textbf{0.064}
        & 0.271& \textbf{0.041}& \textbf{0.087}
        & 0.755& \textbf{0.029}& \textbf{0.101}\\
        \midrule
        \multirow{6}{*}{\texttt{OpenHermes}} & \texttt{Mistral-7B} + SFT $(\beta=0)$ 
        & 0.602 & 0.071 & 0.094 
        & 0.546& 0.041& 0.071
        & 0.867& 0.066& 0.100\\
        & \texttt{Mistral-7B} + SFT $(\beta=0.1)$ 
        & 0.602 & \textbf{0.014} & \textbf{0.059}
        & 0.552& \textbf{0.021}& \textbf{0.042}
        & 0.857& \textbf{0.036}& \textbf{0.076}\\
        \cmidrule{2-11}
        & \texttt{LLaMA3-8B} + SFT $(\beta=0)$ 
        & 0.654 & 0.038 & 0.077
        & 0.552& 0.063& 0.074
        & 0.880& 0.065& 0.112\\
        & \texttt{LLaMA3-8B} + SFT $(\beta=0.1)$ 
        & 0.646 & \textbf{0.016} & \textbf{0.059} 
        & 0.554& \textbf{0.038}& \textbf{0.037}
        & 0.880& \textbf{0.041}& \textbf{0.089}\\
        \cmidrule{2-11}
        & \texttt{Gemma2-2B} + SFT $(\beta=0)$ 
        & 0.541 &0.353 & 0.180  
        & 0.364& 0.125&0.143
        & 0.816& 0.131& 0.175\\
        & \texttt{Gemma2-2B} + SFT $(\beta=0.1)$ 
        & 0.542 & \textbf{0.016} & \textbf{0.063}
        & 0.362& \textbf{0.077}& \textbf{0.096}
        & 0.813& \textbf{0.038}& \textbf{0.096}\\
        \bottomrule
        \end{tabular}
    }
\end{table*}

Our forward pass (\cref{alg:combined}) fuses the computation of all three components of~\cref{eq:ls-breakdown} to minimize memory and indexing costs, rendering the computation efficient on modern GPUs~\citep{cutlass}. First, the embeddings $\mE$ and classifier $\mC$ are divided into chunks $\mE_n$ of $\mE$ with size $D\times N_B$ and $\mC_m$ of $\mC$ with size $D\times V_B$, operated on independently. The standard output $\mO=\mC^\top\mE\in\mathbb{R}^{\lrbracknorm{\mV}\times N}$ is divided into blocks of size $V_B\times N_B$ which store the products $\mO_{nv}=\mC_m^\top\mE_n$. $\mC_m$ and $\mE_n$ are further split along the $D$ dimension into $d$ chunks of size $D_B$. Thus chunks $\mE_{n,d}$ and $\mC_{m,d}$ of size $D_B\times N_B$ and $D_B\times V_B$ can be used to accumulate $\mO_{nv}=\sum_{d}\mC_{m,d}^\top\mE_{n,d}$ directly in SRAM before being written into global memory.

To compute $\mthree$ or the $\mLSE$, the above strategy is sufficient by having each block first compute a matrix multiplication, then the log-sum-exp along the vocabulary dimension $m$ for its block, and finally update the $\mLSE$ with this result. We make use of a trick introduced by \citet{CCE_loss}, where blocks along the same $N$ dimension range but different $\lrbracknorm{\mV}$ dimension range are written in the same location, in order to reduce memory usage on SRAM. This is implemented directly using a lock mechanism, where blocks exchange a single lock per group and update the $\mLSE$ online.

Similar to the above, if the label smoothing parameter $\beta$ is not 0 (indicating that smoothing is used), $\mtwo$ in \cref{eq:ls-breakdown} can be computed through reuse of intermediate results for $\mthree$, by summing over the vocabulary dimension $m$ of $\mO_{nm}$ and updating an output $\vo\in\mathbb{R}^N$ with the result.

To understand how to fuse the computation of $\mone$ into this matrix multiplication, we can first consider that the input sequence $\vx$ can also be split into chunks $\vx_n$ of size $N_B$. Because $\vx_n$ contains information about the target labels, $\vx_n$ can be directly used to compute the memory addresses of the target classifier chunk $\mC_{\vx_i, d}$. Because content from an address for each $\mC_{m,d}$ must be loaded into SRAM to compute $\mO_{nv}$, this means that a direct address comparison can be used to create a mask $\mM\in\mathbb{R}^{N_B\times V_B}$, which has value 1 only where the rows in $\mC_{\vx_i,d}$ matches $\mC_{m,d}$ and 0 otherwise. This enables us to add $\mone$ to $\vo$ by only adding the label-corresponding rows to the loss
\begin{align}
    \vo_n \mathrel{+}=(1-\beta)\sum_m\left[\left(\mE_{n,d}^\top\mC_{m,d}\right)\odot\mM\right],
\end{align}
where $\odot$ is the element-wise matrix multiplication.

Thus for inference, we can directly compute the entire loss efficiently without directly needing to materialize the complete logit matrix in global memory.

\subsubsection{Backward Pass}

To implement a backward pass, borrowing logic from \citet{CCE_loss} is sufficient to produce a backward pass that is significantly faster than existing alternatives. As \citet{CCE_loss} do not consider label smoothing, their backward implementation requires some modification. As the gradient with respect to $\mLSE$ is
\begin{equation}
    \begin{split}
        \nabla_{\mathcal{L}^\text{LS}}^{\mLSE}\mE 
            %&= \nabla_{\mathcal{L}^\text{LS}}\mLSE\frac{\partial}{\partial \mE}\log\sum\exp(\mC^\top\mE)\\
            &=\left((\mathtt{softmax}(\mC^\top\mE)\cdot\nabla_{\mathcal{L}^\text{LS}}\mLSE)\mC\right)^\top,
            % \\&= \mC^\top(\mS \cdot \nabla_{\mathcal{L}^\text{LS}}\mLSE)^\top,
            \\
        \nabla_{\mathcal{L}^\text{LS}}^{\mLSE}\mC 
            %&= \nabla_{\mathcal{L}^\text{LS}}\mLSE\frac{\partial}{\partial \mC}\log\sum\exp(\mC^\top\mE)\\
            &=\left((\mathtt{softmax}(\mC^\top\mE)\cdot\nabla_{\mathcal{L}^\text{LS}}\mLSE)^\top\mE\right)^\top,
            % \\&= \mE^\top(\mS \cdot \nabla_{\mathcal{L}^\text{LS}}\mLSE),
    \end{split}
\end{equation}
incorporating the derivative with respect to $\mone/\mtwo$ only requires adding a constant to $\mathtt{softmax}(\mC^\top\mE)$ depending on whether the element is the true label or not.

While the forward and backward pass can be further fused to compute both the output and gradient simultaneously, this requires computing $\mLSE_{n}$ before any \texttt{softmax} value $\mS_{nv}$, otherwise the computed $\mS_{nv}$ will use an incomplete log-sum-exp scaling factor (each row $\mS_n$ depends on $\mLSE_n$ to have been computed in its entirety). This thus requires a barrier to block subsequent execution code until all preceding computations have been completed by all workers. While can save memory, we observe that specific tricks such as average logit sorting and gradient filtering, introduced by \citet{CCE_loss}, enable faster computation through standalone forward and backward passes with minimal increased memory utilization (2MB overhead).

\subsection{Experiments}

\paragraph{Setup.}
For all models trained on the \texttt{Alpaca}~\citep{alpaca}, \texttt{Tulu3Mixture}~\citep{tulu} and \texttt{OpenHermes}~\citep{OpenHermes} SFT datasets, we adhere to the recommended training configuration outlined by \citet{tulu}. We employ the AdamW optimizer for training and conduct a grid search over the learning rates $\mathtt{\{5e-6, 2e-5, 5e-5, 2e-4\}}$ to determine the optimal setting for each model. To facilitate stable training and prevent over-fitting, we use a batch size of 128 and apply a dropout rate of 0.1.

\paragraph{Calibration Results.} \cref{tab:main_performance} provides a comprehensive comparison of the accuracy and calibration performance of various large language models (LLMs) with and without label smoothing (LS) across different supervised fine-tuning (SFT) datasets. The evaluation is conducted on three widely used benchmark datasets: MMLU, HellaSwag, and ARC-Easy, ensuring a robust assessment of model performance. Across all experimental settings, applying LS with a smoothing factor of $\beta = 0.1$ consistently leads to improved calibration, as indicated by lower Expected Calibration Error (ECE) and Root Mean Square (RMS) calibration error, while preserving model accuracy. Notably, \texttt{LLaMA3-8B} and \texttt{Mistral-7B} achieve the best calibration results, particularly when trained on \texttt{OpenHermes} and \texttt{Tulu3Mixture}. These findings underscore the effectiveness of LS as a simple yet powerful technique for enhancing model reliability without sacrificing predictive accuracy.

\subsection{Benchmarking and Testing}

\paragraph{Benchmarking.} \cref{tab:perf-main}/\ref{tab:perf} provides the primary comparison of our custom kernel against existing alternatives, both with and without label smoothing. We compare in terms of both memory and time.
First, in terms of memory, measured as the peak amount of GPU storage necessary for the computation, our kernels surpass other options, beating the next closest competitor (Liger Kernels) by requiring over 75\% less memory. Time-wise, we trail a compiled \texttt{torch} implementation by less than 10~ms while utilizing only $\approx$7\% the amount of memory as they use.
Furthermore, our method can be applied in more general settings. Under such circumstances, we have two observations. First, we can match Cut-Cross-Entropy (CCE) with a very minimal increase in computation speed (less than 0.5\% overhead for a forward pass and less than 1\% for a combined forward/backward). Furthermore, while an efficient compiled \texttt{torch} implementation remains slightly more efficient, label smoothing causes an increase in running time as well as memory consumption for this specific implementation, factors that are not observed with our method. Accordingly, we can conclude that overall, our method provides a more robust and efficient alternative.

\begin{table}[t!]
    \centering
    \caption{Memory and time to compute losses and gradients. Results are computed on a batch size of 8192 tokens in a single sequence, generated from a \texttt{Gemma2-2B} model (vocabulary size of 256$\mathsf{K}$ and hidden size 2304). Experiments are conducted using \texttt{PyTorch} 2.4.0 and \texttt{CUDA} 12.1. Further see \cref{tab:perf} in \cref{app:additional-results}.}
    \setlength{\tabcolsep}{2pt}
    \resizebox{\linewidth}{!}{
    \begin{tabular}{
        lc rc rc rc rc rc r}
    \toprule
    \multirow{2}{*}{\textbf{Method}} && \multicolumn{3}{c}{\texttt{fwd}} &&  \multicolumn{3}{c}{\texttt{bwd}} && \multicolumn{3}{c}{\texttt{fwd+bwd}} \\
    \cmidrule{3-5}
    \cmidrule{7-9} \cmidrule{11-13}
    && Memory && Time  && Memory  && Time && Memory && Time  \\
    \midrule
    % Lower Bound 
    %     && \qty{0.004}{MB} && 
    %     && \qty{1161}{MB} &&  
    %     && \qty{1161}{MB} && \\
    % \midrule
    \multicolumn{13}{c}{\textbf{Smoothing $\beta > 0$}} \\
    \midrule
    Ours 
        && \qty{1.1}{MB} && \qty{24.2}{ms}
        && \qty{1163}{MB} && \qty{49.3}{ms} 
        && \qty{1164}{MB} && \qty{72.9}{ms} \\
    \midrule
    % \texttt{torchtune}~\citep{torchtune} (8 chunks) 
    %     && \qty{8000}{MB} && \qty{}{ms} 
    %     && \qty{1630}{MB} && \qty{}{ms} 
    %     && \qty{9631}{MB} && \qty{}{ms} \\
    \texttt{torch.compile} 
        && \qty{4000}{MB} && \qty{22.8}{ms} 
        && \qty{12000}{MB} && \qty{38.3}{ms} 
        && \qty{16000}{MB} && \qty{62.3}{ms} \\
    Baseline
        && \qty{24000}{MB} && \qty{41.4}{ms} 
        && \qty{16000}{MB} && \qty{62.5}{ms} 
        && \qty{28000}{MB} && \qty{104.9}{ms} \\
    \midrule
    \multicolumn{13}{c}{\textbf{Smoothing $\beta = 0$}} \\
    \midrule
    Ours 
        && \qty{1.1}{MB} && \qty{24.0}{ms} 
        && \qty{1163}{MB} && \qty{49.2}{ms} 
        && \qty{1164}{MB} && \qty{72.9}{ms}\\
    \midrule
    Cut-Cross Entropy\tablefootnote{This method does not support label smoothing.} 
        && \qty{1.1}{MB} && \qty{23.6}{ms}
        && \qty{1163}{MB} && \qty{49.2}{ms} 
        && \qty{1164}{MB} && \qty{72.4}{ms} \\
    \texttt{torch.compile}
        && \qty{4000}{MB} && \qty{20.6}{ms} 
        && \qty{4000}{MB} && \qty{33.9}{ms}
        && \qty{8000}{MB} && \qty{55.0}{ms} \\
    Baseline
        && \qty{24000}{MB} && \qty{38.7}{ms} 
        && \qty{16000}{MB} && \qty{55.8}{ms} 
        && \qty{28000}{MB} && \qty{96.0}{ms} \\
    \bottomrule
    \end{tabular}}
    \label{tab:perf-main}
    \gdef\rownumber{}
\end{table}

\begin{figure}[th]
    \centering
    % \vspace{-0.5cm}
    \includegraphics[width=0.9\linewidth]{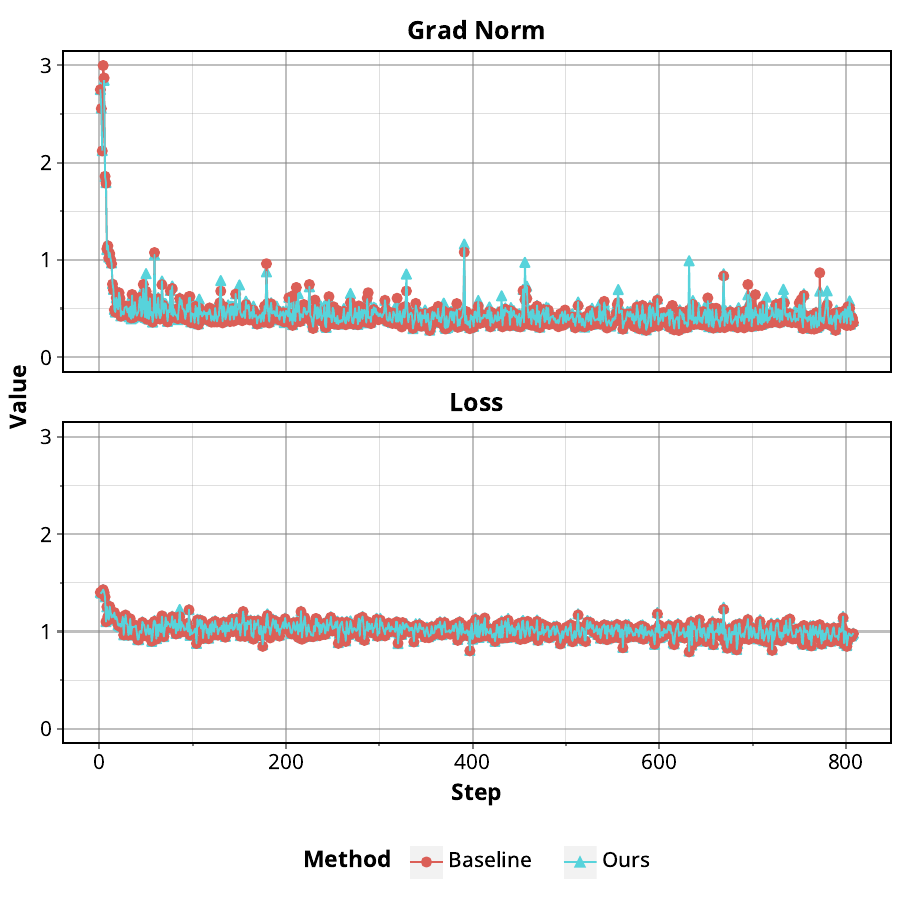}
    % \missingfigure[figwidth=\linewidth]{Training curves}
    \vspace{-0.25cm}
    \caption{Training curves comparing our implementation against \texttt{torch.nn.CrossEntropyLoss} with $\beta = 0.1$. Training uses a \texttt{LLaMA3.2-3B} model using the \texttt{Alpaca} dataset.}
    \label{fig:training-cures}
\end{figure}

\paragraph{Training Results.} We further compare models trained with our custom kernel and to those using a baseline/reference \texttt{torch} implementation in \cref{fig:training-cures}. We demonstrate here that both methods are indistinguishable in terms of loss curves as well as gradient norms, indicating the correctness and stability of the kernel.

\section{Conclusion}

In this paper, we present a novel perspective regarding the use of label smoothing as a functional mechanism to retain calibration after the supervised fine-tuning process (SFT) that is commonly used to train LLMs. We first identify why SFT can degrade calibration from a statistical standpoint before showing how label smoothing can help alleviate this concern. However, we also identify a setting where the use of label smoothing appears ineffective, particularly in the case of large language models with large vocabularies but smaller hidden sizes, demonstrating how such model construction implicitly impacts the entropy of predictions and leads to difficulty in becoming over-confident during SFT. Finally, we demonstrate a practical issue regarding the computational concerns of label smoothing in these settings. 
To further address this issue, we identify GPU accelerator level optimizations and provide a custom computational kernel, written in Triton, that enables us to maximize the use of accelerator memory/bandwidth and train models, improving both training/inference speed as well as memory consumption without sacrifices in stability.

\icmlack{
    Jerry Huang was supported by the NSERC Canada Graduate Scholarships — Doctoral (CGS-D) program (funding reference number 589326) as well as the Bourse d'\'{E}xcellence Hydro-Qu\'{e}bec program. This work was made possible in part thanks to computational resources from Calcul Qu\'{e}bec\footnote{\url{https://www.calculquebec.ca/}} and the Digital Research Alliance of Canada (DRAC)\footnote{\url{https://alliancecan.ca/en}}.
}

\section*{Impact Statement}

This paper proposes a method to improve calibration in large-vocabulary language models. We anticipate minimal societal impact from this work.

\bibliography{ref}

\begin{thebibliography}{49}
\providecommand{\natexlab}[1]{#1}
\providecommand{\url}[1]{\texttt{#1}}
\expandafter\ifx\csname urlstyle\endcsname\relax
  \providecommand{\doi}[1]{doi: #1}\else
  \providecommand{\doi}{doi: \begingroup \urlstyle{rm}\Url}\fi

\bibitem[Bach et~al.(2022)Bach, Sanh, Yong, Webson, Raffel, Nayak, Sharma, Kim, Bari, F{\'{e}}vry, Alyafeai, Dey, Santilli, Sun, Ben{-}David, Xu, Chhablani, Wang, Fries, AlShaibani, Sharma, Thakker, Almubarak, Tang, Radev, Jiang, and Rush]{promptsource}
Bach, S.~H., Sanh, V., Yong, Z.~X., Webson, A., Raffel, C., Nayak, N.~V., Sharma, A., Kim, T., Bari, M.~S., F{\'{e}}vry, T., Alyafeai, Z., Dey, M., Santilli, A., Sun, Z., Ben{-}David, S., Xu, C., Chhablani, G., Wang, H., Fries, J.~A., AlShaibani, M.~S., Sharma, S., Thakker, U., Almubarak, K., Tang, X., Radev, D.~R., Jiang, M.~T., and Rush, A.~M.
\newblock Promptsource: An integrated development environment and repository for natural language prompts.
\newblock In \emph{Proceedings of the 60th Annual Meeting of the Association for Computational Linguistics, {ACL} 2022 - System Demonstrations, Dublin, Ireland, May 22-27, 2022}, pp.\  93--104, 2022.

\bibitem[Bertsekas(1999)]{nonlinear_programming}
Bertsekas, D.~P.
\newblock \emph{Nonlinear Programming}.
\newblock 1999.

\bibitem[Brier(1950)]{statistical_calibration}
Brier, G.~W.
\newblock Verification of forecasts expressed in terms of probability.
\newblock \emph{Monthly Weather Review}, 78\penalty0 (1):\penalty0 1--3, 1950.

\bibitem[Chi et~al.(2024)Chi, Fan, and Rudnicky]{logitsnormal}
Chi, T., Fan, T., and Rudnicky, A.
\newblock Attention alignment and flexible positional embeddings improve transformer length extrapolation.
\newblock In \emph{Findings of the Association for Computational Linguistics: {NAACL} 2024, Mexico City, Mexico, June 16-21, 2024}, pp.\  132--148, 2024.

\bibitem[Chung et~al.(2024)Chung, Hou, Longpre, Zoph, Tay, Fedus, Li, Wang, Dehghani, Brahma, Webson, Gu, Dai, Suzgun, Chen, Chowdhery, Castro{-}Ros, Pellat, Robinson, Valter, Narang, Mishra, Yu, Zhao, Huang, Dai, Yu, Petrov, Chi, Dean, Devlin, Roberts, Zhou, Le, and Wei]{it_scaling}
Chung, H.~W., Hou, L., Longpre, S., Zoph, B., Tay, Y., Fedus, W., Li, Y., Wang, X., Dehghani, M., Brahma, S., Webson, A., Gu, S.~S., Dai, Z., Suzgun, M., Chen, X., Chowdhery, A., Castro{-}Ros, A., Pellat, M., Robinson, K., Valter, D., Narang, S., Mishra, G., Yu, A., Zhao, V.~Y., Huang, Y., Dai, A.~M., Yu, H., Petrov, S., Chi, E.~H., Dean, J., Devlin, J., Roberts, A., Zhou, D., Le, Q.~V., and Wei, J.
\newblock Scaling instruction-finetuned language models.
\newblock \emph{Journal on Machine Learning Research}, 25:\penalty0 70:1--70:53, 2024.

\bibitem[Dao(2024)]{fla2}
Dao, T.
\newblock Flashattention-2: Faster attention with better parallelism and work partitioning.
\newblock In \emph{The Twelfth International Conference on Learning Representations, {ICLR} 2024, Vienna, Austria, May 7-11, 2024}, 2024.

\bibitem[Dao et~al.(2022)Dao, Fu, Ermon, Rudra, and R{\'{e}}]{fla}
Dao, T., Fu, D.~Y., Ermon, S., Rudra, A., and R{\'{e}}, C.
\newblock Flashattention: Fast and memory-efficient exact attention with io-awareness.
\newblock In \emph{Advances in Neural Information Processing Systems 35: Annual Conference on Neural Information Processing Systems 2022, NeurIPS 2022, New Orleans, LA, USA, November 28 - December 9, 2022}, 2022.

\bibitem[DeGroot \& Fienberg(1983)DeGroot and Fienberg]{rss1983}
DeGroot, M.~H. and Fienberg, S.~E.
\newblock The comparison and evaluation of forecasters.
\newblock \emph{Journal of the Royal Statistical Society. Series D (The Statistician)}, 32\penalty0 (1):\penalty0 12--22, 1983.

\bibitem[Dubois et~al.(2023)Dubois, Li, Taori, Zhang, Gulrajani, Ba, Guestrin, Liang, and Hashimoto]{alpaca}
Dubois, Y., Li, C.~X., Taori, R., Zhang, T., Gulrajani, I., Ba, J., Guestrin, C., Liang, P., and Hashimoto, T.~B.
\newblock Alpacafarm: {A} simulation framework for methods that learn from human feedback.
\newblock In \emph{Advances in Neural Information Processing Systems 36: Annual Conference on Neural Information Processing Systems 2023, NeurIPS 2023, New Orleans, LA, USA, December 10 - 16, 2023}, 2023.

\bibitem[{Gemma Team}(2024)]{gemma2}
{Gemma Team}.
\newblock Gemma 2: Improving open language models at a practical size, 2024.

\bibitem[Grattafiori et~al.(2024)]{llama3}
Grattafiori, A. et~al.
\newblock The llama 3 herd of models, 2024.

\bibitem[Guo et~al.(2017)Guo, Pleiss, Sun, and Weinberger]{Calibration_NN}
Guo, C., Pleiss, G., Sun, Y., and Weinberger, K.~Q.
\newblock On calibration of modern neural networks.
\newblock In \emph{Proceedings of the 34th International Conference on Machine Learning, {ICML} 2017, Sydney, NSW, Australia, 6-11 August 2017}, volume~70 of \emph{Proceedings of Machine Learning Research}, pp.\  1321--1330, 2017.

\bibitem[Hendrycks et~al.(2019)Hendrycks, Mazeika, and Dietterich]{RMS_CE}
Hendrycks, D., Mazeika, M., and Dietterich, T.~G.
\newblock Deep anomaly detection with outlier exposure.
\newblock In \emph{7th International Conference on Learning Representations, {ICLR} 2019, New Orleans, LA, USA, May 6-9, 2019}, 2019.

\bibitem[Hendrycks et~al.(2021)Hendrycks, Burns, Basart, Zou, Mazeika, Song, and Steinhardt]{mmlu}
Hendrycks, D., Burns, C., Basart, S., Zou, A., Mazeika, M., Song, D., and Steinhardt, J.
\newblock Measuring massive multitask language understanding.
\newblock In \emph{9th International Conference on Learning Representations, {ICLR} 2021, Virtual Event, Austria, May 3-7, 2021}, 2021.

\bibitem[Hsu et~al.(2024)Hsu, Dai, Kothapalli, Song, Tang, Zhu, Shimizu, Sahni, Ning, and Chen]{liger}
Hsu, P.-L., Dai, Y., Kothapalli, V., Song, Q., Tang, S., Zhu, S., Shimizu, S., Sahni, S., Ning, H., and Chen, Y.
\newblock Liger kernel: Efficient triton kernels for llm training, 2024.

\bibitem[Huh et~al.(2024)Huh, Cheung, Wang, and Isola]{platonic_representations}
Huh, M., Cheung, B., Wang, T., and Isola, P.
\newblock Position: The platonic representation hypothesis.
\newblock In \emph{Forty-first International Conference on Machine Learning, {ICML} 2024, Vienna, Austria, July 21-27, 2024}, 2024.

\bibitem[Jiang et~al.(2023)Jiang, Sablayrolles, Mensch, Bamford, Chaplot, de~Las~Casas, Bressand, Lengyel, Lample, Saulnier, Lavaud, Lachaux, Stock, Scao, Lavril, Wang, Lacroix, and Sayed]{mistral}
Jiang, A.~Q., Sablayrolles, A., Mensch, A., Bamford, C., Chaplot, D.~S., de~Las~Casas, D., Bressand, F., Lengyel, G., Lample, G., Saulnier, L., Lavaud, L.~R., Lachaux, M., Stock, P., Scao, T.~L., Lavril, T., Wang, T., Lacroix, T., and Sayed, W.~E.
\newblock Mistral 7b, 2023.

\bibitem[Jiang et~al.(2021)Jiang, Araki, Ding, and Neubig]{calibration_qa}
Jiang, Z., Araki, J., Ding, H., and Neubig, G.
\newblock How can we know \emph{When} language models know? on the calibration of language models for question answering.
\newblock \emph{Trans. Assoc. Comput. Linguistics}, 9:\penalty0 962--977, 2021.

\bibitem[Kerr et~al.(2017)Kerr, Merrill, Demouth, and Tran]{cutlass}
Kerr, A., Merrill, D., Demouth, J., and Tran, J.
\newblock Cutlass: Fast linear algebra in cuda c++, December 2017.

\bibitem[Kumar et~al.(2022)Kumar, Raghunathan, Jones, Ma, and Liang]{finetuning_distort}
Kumar, A., Raghunathan, A., Jones, R.~M., Ma, T., and Liang, P.
\newblock Fine-tuning can distort pretrained features and underperform out-of-distribution.
\newblock In \emph{The Tenth International Conference on Learning Representations, {ICLR} 2022, Virtual Event, April 25-29, 2022}, 2022.

\bibitem[Lin et~al.(2017)Lin, Goyal, Girshick, He, and Doll{\'{a}}r]{focal_loss}
Lin, T., Goyal, P., Girshick, R.~B., He, K., and Doll{\'{a}}r, P.
\newblock Focal loss for dense object detection.
\newblock In \emph{{IEEE} International Conference on Computer Vision, {ICCV} 2017, Venice, Italy, October 22-29, 2017}, pp.\  2999--3007, 2017.

\bibitem[Liu et~al.(2022)Liu, Ayed, Galdran, and Dolz]{margin_based_label_smoothing}
Liu, B., Ayed, I.~B., Galdran, A., and Dolz, J.
\newblock The devil is in the margin: Margin-based label smoothing for network calibration.
\newblock In \emph{{IEEE/CVF} Conference on Computer Vision and Pattern Recognition, {CVPR} 2022, New Orleans, LA, USA, June 18-24, 2022}, pp.\  80--88, 2022.

\bibitem[Lu et~al.(2023)Lu, Rashid, Kobyzev, Rezagholizadeh, and Langlais]{LABO}
Lu, P., Rashid, A., Kobyzev, I., Rezagholizadeh, M., and Langlais, P.
\newblock {LABO:} towards learning optimal label regularization via bi-level optimization.
\newblock In \emph{Findings of the Association for Computational Linguistics: {ACL} 2023, Toronto, Canada, July 9-14, 2023}, pp.\  5759--5774, 2023.

\bibitem[Lukasik et~al.(2020)Lukasik, Bhojanapalli, Menon, and Kumar]{smoothing_noisy_labels}
Lukasik, M., Bhojanapalli, S., Menon, A.~K., and Kumar, S.
\newblock Does label smoothing mitigate label noise?
\newblock In \emph{Proceedings of the 37th International Conference on Machine Learning, {ICML} 2020, 13-18 July 2020, Virtual Event}, volume 119 of \emph{Proceedings of Machine Learning Research}, pp.\  6448--6458, 2020.

\bibitem[Minderer et~al.(2021)Minderer, Djolonga, Romijnders, Hubis, Zhai, Houlsby, Tran, and Lucic]{revisiting_calibration}
Minderer, M., Djolonga, J., Romijnders, R., Hubis, F., Zhai, X., Houlsby, N., Tran, D., and Lucic, M.
\newblock Revisiting the calibration of modern neural networks.
\newblock In \emph{Advances in Neural Information Processing Systems 34: Annual Conference on Neural Information Processing Systems 2021, NeurIPS 2021, December 6-14, 2021, virtual}, pp.\  15682--15694, 2021.

\bibitem[Mukhoti et~al.(2020)Mukhoti, Kulharia, Sanyal, Golodetz, Torr, and Dokania]{focal_loss_calibration}
Mukhoti, J., Kulharia, V., Sanyal, A., Golodetz, S., Torr, P. H.~S., and Dokania, P.~K.
\newblock Calibrating deep neural networks using focal loss.
\newblock In \emph{Advances in Neural Information Processing Systems 33: Annual Conference on Neural Information Processing Systems 2020, NeurIPS 2020, December 6-12, 2020, virtual}, 2020.

\bibitem[Mukhoti et~al.(2024)Mukhoti, Gal, Torr, and Dokania]{finetuning_cripple}
Mukhoti, J., Gal, Y., Torr, P., and Dokania, P.~K.
\newblock Fine-tuning can cripple your foundation model; preserving features may be the solution.
\newblock \emph{Transactions on Machine Learning Research}, 2024, 2024.

\bibitem[M{\"{u}}ller et~al.(2019)M{\"{u}}ller, Kornblith, and Hinton]{when_ls_work}
M{\"{u}}ller, R., Kornblith, S., and Hinton, G.~E.
\newblock When does label smoothing help?
\newblock In \emph{Advances in Neural Information Processing Systems 32: Annual Conference on Neural Information Processing Systems 2019, NeurIPS 2019, December 8-14, 2019, Vancouver, BC, Canada}, pp.\  4696--4705, 2019.

\bibitem[Murphy(1972)]{murphy1972}
Murphy, A.~H.
\newblock Scalar and vector partitions of the probability score: Part i. two-state situation.
\newblock \emph{Journal of Applied Meteorology and Climatology}, 11\penalty0 (2):\penalty0 273--282, 1972.

\bibitem[Naeini et~al.(2015)Naeini, Cooper, and Hauskrecht]{ECE}
Naeini, M.~P., Cooper, G.~F., and Hauskrecht, M.
\newblock Obtaining well calibrated probabilities using bayesian binning.
\newblock In \emph{Proceedings of the Twenty-Ninth {AAAI} Conference on Artificial Intelligence, January 25-30, 2015, Austin, Texas, {USA}}, pp.\  2901--2907, 2015.

\bibitem[Nixon et~al.(2019)Nixon, Dusenberry, Zhang, Jerfel, and Tran]{SCE_ACE}
Nixon, J., Dusenberry, M.~W., Zhang, L., Jerfel, G., and Tran, D.
\newblock Measuring calibration in deep learning.
\newblock In \emph{{IEEE} Conference on Computer Vision and Pattern Recognition Workshops, {CVPR} Workshops 2019, Long Beach, CA, USA, June 16-20, 2019}, pp.\  38--41, 2019.

\bibitem[Oh et~al.(2024)Oh, Lim, Kim, Han, Yun, Choo, Hauptmann, Cheng, and Song]{calibrated_finetuning}
Oh, C., Lim, H., Kim, M., Han, D., Yun, S., Choo, J., Hauptmann, A.~G., Cheng, Z.-Q., and Song, K.
\newblock Towards calibrated robust fine-tuning of vision-language models.
\newblock In \emph{Advances in Neural Information Processing Systems 38: Annual Conference on Neural Information Processing Systems 2024, NeurIPS 2024, December 10-15, 2024, Vancouver, BC, Canada}, 2024.

\bibitem[Ouyang et~al.(2022)Ouyang, Wu, Jiang, Almeida, Wainwright, Mishkin, Zhang, Agarwal, Slama, Ray, Schulman, Hilton, Kelton, Miller, Simens, Askell, Welinder, Christiano, Leike, and Lowe]{it_human_feedback}
Ouyang, L., Wu, J., Jiang, X., Almeida, D., Wainwright, C.~L., Mishkin, P., Zhang, C., Agarwal, S., Slama, K., Ray, A., Schulman, J., Hilton, J., Kelton, F., Miller, L., Simens, M., Askell, A., Welinder, P., Christiano, P.~F., Leike, J., and Lowe, R.
\newblock Training language models to follow instructions with human feedback.
\newblock In \emph{Advances in Neural Information Processing Systems 35: Annual Conference on Neural Information Processing Systems 2022, NeurIPS 2022, New Orleans, LA, USA, November 28 - December 9, 2022}, 2022.

\bibitem[Pereyra et~al.(2017)Pereyra, Tucker, Chorowski, Kaiser, and Hinton]{penalizing_confidence}
Pereyra, G., Tucker, G., Chorowski, J., Kaiser, L., and Hinton, G.~E.
\newblock Regularizing neural networks by penalizing confident output distributions.
\newblock In \emph{5th International Conference on Learning Representations, {ICLR} 2017, Toulon, France, April 24-26, 2017, Workshop Track Proceedings}, 2017.

\bibitem[{PyTorch}(2024)]{torchtune}
{PyTorch}.
\newblock torchtune: Pytorch's finetuning library, April 2024.

\bibitem[Sanh et~al.(2022)Sanh, Webson, Raffel, Bach, Sutawika, Alyafeai, Chaffin, Stiegler, Raja, Dey, Bari, Xu, Thakker, Sharma, Szczechla, Kim, Chhablani, Nayak, Datta, Chang, Jiang, Wang, Manica, Shen, Yong, Pandey, Bawden, Wang, Neeraj, Rozen, Sharma, Santilli, F{\'{e}}vry, Fries, Teehan, Scao, Biderman, Gao, Wolf, and Rush]{prompting_zsg}
Sanh, V., Webson, A., Raffel, C., Bach, S.~H., Sutawika, L., Alyafeai, Z., Chaffin, A., Stiegler, A., Raja, A., Dey, M., Bari, M.~S., Xu, C., Thakker, U., Sharma, S.~S., Szczechla, E., Kim, T., Chhablani, G., Nayak, N.~V., Datta, D., Chang, J., Jiang, M.~T., Wang, H., Manica, M., Shen, S., Yong, Z.~X., Pandey, H., Bawden, R., Wang, T., Neeraj, T., Rozen, J., Sharma, A., Santilli, A., F{\'{e}}vry, T., Fries, J.~A., Teehan, R., Scao, T.~L., Biderman, S., Gao, L., Wolf, T., and Rush, A.~M.
\newblock Multitask prompted training enables zero-shot task generalization.
\newblock In \emph{The Tenth International Conference on Learning Representations, {ICLR} 2022, Virtual Event, April 25-29, 2022}, 2022.

\bibitem[Szegedy et~al.(2016)Szegedy, Vanhoucke, Ioffe, Shlens, and Wojna]{inception}
Szegedy, C., Vanhoucke, V., Ioffe, S., Shlens, J., and Wojna, Z.
\newblock Rethinking the inception architecture for computer vision.
\newblock In \emph{2016 {IEEE} Conference on Computer Vision and Pattern Recognition, {CVPR} 2016, Las Vegas, NV, USA, June 27-30, 2016}, pp.\  2818--2826, 2016.

\bibitem[Teknium(2023)]{OpenHermes}
Teknium.
\newblock Openhermes 2.5: An open dataset of synthetic data for generalist llm assistants, 2023.

\bibitem[Wang et~al.(2022)Wang, Mishra, Alipoormolabashi, Kordi, Mirzaei, Naik, Ashok, Dhanasekaran, Arunkumar, Stap, Pathak, Karamanolakis, Lai, Purohit, Mondal, Anderson, Kuznia, Doshi, Pal, Patel, Moradshahi, Parmar, Purohit, Varshney, Kaza, Verma, Puri, Karia, Doshi, Sampat, Mishra, A, Patro, Dixit, and Shen]{supernli}
Wang, Y., Mishra, S., Alipoormolabashi, P., Kordi, Y., Mirzaei, A., Naik, A., Ashok, A., Dhanasekaran, A.~S., Arunkumar, A., Stap, D., Pathak, E., Karamanolakis, G., Lai, H.~G., Purohit, I., Mondal, I., Anderson, J., Kuznia, K., Doshi, K., Pal, K.~K., Patel, M., Moradshahi, M., Parmar, M., Purohit, M., Varshney, N., Kaza, P.~R., Verma, P., Puri, R.~S., Karia, R., Doshi, S., Sampat, S.~K., Mishra, S., A, S.~R., Patro, S., Dixit, T., and Shen, X.
\newblock Super-naturalinstructions: Generalization via declarative instructions on 1600+ {NLP} tasks.
\newblock In \emph{Proceedings of the 2022 Conference on Empirical Methods in Natural Language Processing, {EMNLP} 2022, Abu Dhabi, United Arab Emirates, December 7-11, 2022}, pp.\  5085--5109, 2022.

\bibitem[Wang et~al.(2023)Wang, Ivison, Dasigi, Hessel, Khot, Chandu, Wadden, MacMillan, Smith, Beltagy, and Hajishirzi]{tulu}
Wang, Y., Ivison, H., Dasigi, P., Hessel, J., Khot, T., Chandu, K.~R., Wadden, D., MacMillan, K., Smith, N.~A., Beltagy, I., and Hajishirzi, H.
\newblock How far can camels go? exploring the state of instruction tuning on open resources.
\newblock In \emph{Advances in Neural Information Processing Systems 36: Annual Conference on Neural Information Processing Systems 2023, NeurIPS 2023, New Orleans, LA, USA, December 10 - 16, 2023}, 2023.

\bibitem[Wei et~al.(2022{\natexlab{a}})Wei, Bosma, Zhao, Guu, Yu, Lester, Du, Dai, and Le]{sft_models_zero_shot_learners}
Wei, J., Bosma, M., Zhao, V.~Y., Guu, K., Yu, A.~W., Lester, B., Du, N., Dai, A.~M., and Le, Q.~V.
\newblock Finetuned language models are zero-shot learners.
\newblock In \emph{The Tenth International Conference on Learning Representations, {ICLR} 2022, Virtual Event, April 25-29, 2022}, 2022{\natexlab{a}}.

\bibitem[Wei et~al.(2022{\natexlab{b}})Wei, Liu, Liu, Niu, Sugiyama, and Liu]{smoothing_noisy_labels2}
Wei, J., Liu, H., Liu, T., Niu, G., Sugiyama, M., and Liu, Y.
\newblock To smooth or not? when label smoothing meets noisy labels.
\newblock In \emph{International Conference on Machine Learning, {ICML} 2022, 17-23 July 2022, Baltimore, Maryland, {USA}}, volume 162 of \emph{Proceedings of Machine Learning Research}, pp.\  23589--23614, 2022{\natexlab{b}}.

\bibitem[Wijmans et~al.(2025)Wijmans, Huval, Hertzberg, Koltun, and Kr{\"{a}}henb{\"{u}}hl]{CCE_loss}
Wijmans, E., Huval, B., Hertzberg, A., Koltun, V., and Kr{\"{a}}henb{\"{u}}hl, P.
\newblock Cut your losses in large-vocabulary language models.
\newblock In \emph{The Thirteenth International Conference on Learning Representations, {ICLR} 2025, Singapore, April 24-28, 2025}, 2025.

\bibitem[Xiong et~al.(2024)Xiong, Hu, Lu, Li, Fu, He, and Hooi]{llms_confidence}
Xiong, M., Hu, Z., Lu, X., Li, Y., Fu, J., He, J., and Hooi, B.
\newblock Can llms express their uncertainty? an empirical evaluation of confidence elicitation in llms.
\newblock In \emph{The Twelfth International Conference on Learning Representations, {ICLR} 2024, Vienna, Austria, May 7-11, 2024}, 2024.

\bibitem[Yang et~al.(2024)Yang, Yang, Zhang, Hui, Zheng, Yu, Li, Liu, Huang, Wei, et~al.]{qwen2.5}
Yang, A., Yang, B., Zhang, B., Hui, B., Zheng, B., Yu, B., Li, C., Liu, D., Huang, F., Wei, H., et~al.
\newblock Qwen2.5 technical report, 2024.

\bibitem[Zhai et~al.(2023)Zhai, Likhomanenko, Littwin, Busbridge, Ramapuram, Zhang, Gu, and Susskind]{entropy_attn}
Zhai, S., Likhomanenko, T., Littwin, E., Busbridge, D., Ramapuram, J., Zhang, Y., Gu, J., and Susskind, J.~M.
\newblock Stabilizing transformer training by preventing attention entropy collapse.
\newblock In \emph{International Conference on Machine Learning, {ICML} 2023, 23-29 July 2023, Honolulu, Hawaii, {USA}}, volume 202 of \emph{Proceedings of Machine Learning Research}, pp.\  40770--40803, 2023.

\bibitem[Zhang \& Sabuncu(2020)Zhang and Sabuncu]{ls_selfdistillation}
Zhang, Z. and Sabuncu, M.~R.
\newblock Self-distillation as instance-specific label smoothing.
\newblock In \emph{Advances in Neural Information Processing Systems 33: Annual Conference on Neural Information Processing Systems 2020, NeurIPS 2020, December 6-12, 2020, virtual}, 2020.

\bibitem[Zhao et~al.(2021)Zhao, Wallace, Feng, Klein, and Singh]{contextual_calibration}
Zhao, Z., Wallace, E., Feng, S., Klein, D., and Singh, S.
\newblock Calibrate before use: Improving few-shot performance of language models.
\newblock In \emph{Proceedings of the 38th International Conference on Machine Learning, {ICML} 2021, 18-24 July 2021, Virtual Event}, volume 139 of \emph{Proceedings of Machine Learning Research}, pp.\  12697--12706, 2021.

\bibitem[Łukasz Rajkowski(2019)]{analysisofmap}
Łukasz Rajkowski.
\newblock {Analysis of the Maximal a Posteriori Partition in the Gaussian Dirichlet Process Mixture Model}.
\newblock \emph{Bayesian Analysis}, 14\penalty0 (2):\penalty0 477 -- 494, 2019.

\end{thebibliography}
\bibliographystyle{icml2025}

%%%%%%%%%%%%%%%%%%%%%%%%%%%%%%%%%%%%%%%%%%%%%%%%%%%%%%%%%%%%%%%%%%%%%%%%%%%%%%%
%%%%%%%%%%%%%%%%%%%%%%%%%%%%%%%%%%%%%%%%%%%%%%%%%%%%%%%%%%%%%%%%%%%%%%%%%%%%%%%
% APPENDIX
%%%%%%%%%%%%%%%%%%%%%%%%%%%%%%%%%%%%%%%%%%%%%%%%%%%%%%%%%%%%%%%%%%%%%%%%%%%%%%%
%%%%%%%%%%%%%%%%%%%%%%%%%%%%%%%%%%%%%%%%%%%%%%%%%%%%%%%%%%%%%%%%%%%%%%%%%%%%%%%
\newpage
\appendix
\onecolumn

\section{Proofs}\label{sec:proofs}
\subsection{Proof of \cref{prop:consraint}}\label{sec:constraint_proof}

\textbf{\cref{prop:consraint}}.
    A linear penalty (or a Lagrangian term) for the hard constraint $\bm{d}(\vx) = \bm{0}$ is bounded from above and below by $\mathrm{KL}\left(\vu\|\hat{\vxigma}\left(\vx;\vtheta\right)\right)$, up to additive constants
    \begin{equation}
        \mathrm{KL}[\vu\|\hat{\vxigma}\left(\vx;\vtheta\right)]-\log K \leq \sum_{i=1}^K\bm{d}\left(\vx\right)_i/{K}\leq \mathrm{KL}\left[\vu\|\hat{\vxigma}\left(\vx;\vtheta\right)\right].
    \end{equation}

\begin{proof}
We adapt the proof of \citet{margin_based_label_smoothing}. Given the KL divergence
\[
\mathrm{KL}\left[\vu\|\hat{\vxigma}\left(\vx;\vtheta\right)\right] = -\frac{1}{K}\sum_{k=1}^K\log{P\left(\gamma_i|\vx;\vtheta\right)} + \text{const}
\]
we have that
\begin{equation}
    \begin{split}
        \mathrm{KL}\left[\vu\|\hat{\vxigma}\left(\vx;\vtheta\right)\right] =& -\frac{1}{K}\sum_{k=1}^K\log\left(\frac{e^{\vell\left(\vx;\vtheta\right)_i}}{\sum_{j=1}^Ke^{\vell\left(\vx;\vtheta\right)_j}}\right) + \text{const}\\
        =&-\frac{1}{K}\sum_{k=1}^K\log\left(\sum_{j=1}^Ke^{\vell\left(\vx;\vtheta\right)_j}-\vell\left(\vx;\vtheta\right)_i\right) + \text{const}\\
    \end{split}
\end{equation}
Considering the property of the LogSumExp (LSE) function, it follows that
\[
\max_j\phantom{0}\vell\left(\vx;\vtheta\right)_j\leq \log\sum_{j=1}^K e^{\vell\left(\vx;\vtheta\right)_j}\leq \max_j\phantom{0}\vell\left(\vx;\vtheta\right)_j+\log\left(K\right)
\]
and
\begin{equation}
    \mathrm{KL}\left[\vu\|\hat{\vxigma}\left(\vx;\vtheta\right)\right]-\log K 
    \leq -\frac{1}{K}\sum_{k=1}^K \left(\max_j\phantom{0}\vell\left(\vx;\vtheta\right)_j-\vell\left(\vx;\vtheta\right)_k\right) \\
    \leq \mathrm{KL}\left[\vu\|\hat{\vxigma}\left(\vx;\vtheta\right)\right]
\end{equation}
and given the definition of $\vd\left(\vx\right)$, then the additional penalty $\mathrm{KL}\left[\vu\|\hat{\vxigma}\left(\vx;\vtheta\right)\right]$ imposed by LS in addition to the standard cross-entropy loss $\mathcal{L}^{\text{CE}}$ is approximately optimizing a linear penalty (or a Lagrangian) for the constraint
\[\vd\left(\vx\right)=\bm{0}\] to encourage equality of the logits.
\end{proof}

\subsection{Proof of \cref{prop:ls-map}}\label{sec:proof-ls-map}

\textbf{\cref{prop:ls-map}.} Define a likelihood model $p\left(y | \vx; \vtheta\right) = \mathrm{Cat}\left( \mathrm{softmax}\left(f\left(\vx;\vtheta\right)\right) \right)$, a categorical distribution with parameters $\vz=\mathrm{softmax}\left(f\left(\vx;\vtheta\right)\right)  \in \Delta(\Theta)$ where $\Delta\left(\Theta\right)$ denotes a probability simplex over the parameter space $\Theta$. The label smoothing objective is equivalent to Maximum A Posteriori (MAP) estimation on the $\mathrm{softmax}$ probability vector under the independence assumption $p\left(\vz|\vx\right) = p\left(\vz\right)$.

\begin{proof}
    This proof is an adaptation of \citet{ls_selfdistillation}. Suppose a provide set of examples $\mathcal{D}=\left\{\left(\vx_n,y_n\right)\right\}_{i=1}^N$ sampled from $\mathcal{X}\times\mathcal{Y}$. The goal is to find a set of parameters $\vtheta$ to parameterize a function $f$ that maps inputs $\vx\in \mathcal{X}$ to corresponding labels $y\in\mathcal{Y}$. Suppose that the likelihood $p\left(y | \vx, \vz\right) = \mathtt{Cat}(\vz)$ be a categorical distribution with parameter $\vz \in \Delta\left(\Theta\right)$ and the conditional prior $p\left(\vz | \vx\right) = \mathtt{Dir}\left(\valpha_{\vx}\right)$ be a Dirichlet distribution with instance-specific parameter $\valpha_{\vx}$.
    
    Due to conjugacy of the Dirichlet prior, a closed-form solution of $\hat{\vz}_i = \frac{\mC_i + {\valpha_{\vx_i}} - 1}{ \sum_j \mC_j + {\valpha_{\vx_j}} - 1}$, where $\mC_i$ corresponds to number of occurrences of the $i$-th category, can be easily obtained.
    
    Thus the MAP estimation $\hat{\vz}_i \approx \mathrm{softmax}\left(f_{\vw}\left(\vx_i\right)\right)$ can be amortized with a given training set, resulting in an optimization problem of:
    \begin{align}
        &\max_{\vtheta} \frac{1}{N} \sum_{n=1}^N \log p\left(\vz|\vx_n,y_n;\vtheta, \valpha_x\right) \\
        = &\max_{\vtheta} \sum_{n=1}^N \log p\left(y = y_n|\vz,\vx_n; \vtheta\right) + \log p\left(\vz|\vx_n;\vtheta, \valpha_x\right) \nonumber \\
        % &= \max_{\vw} \frac{1}{n}\sum_{i=1}^n \left( \log \vz^{(y_i)} +  \log \Gamma \left( \sum^k_{c=1} \valpha_{\vx_i}^{(c)} \right) - \sum^k_{c=1} \log \Gamma \left( \valpha_{\vx_i}^{(c)} \right) + \sum^k_{c=1} (\valpha_{\vx_i}^{(c)} - 1) \log \vz^{(c)} \right)\\
        = &\max_{\vw} \underbrace{\frac{1}{N}\sum_{n=1}^N \log \left[\mathrm{softmax}\left(f\left(\vx_n;\vtheta\right)\right)\right]_{y_n}}_\text{Cross Entropy} + 
        \underbrace{\frac{1}{N}\sum_{n=1}^N \sum^K_{k=1} \left([\valpha_{\vx_n}]_{k} - 1\right) \log [\vz]_{k}}_\text{Instance-specific Regularization}
        \label{eq:MAP1}
    \end{align}
    where $[\cdot]_k$ denotes the $k$-th element of a vector.
    Using the assumption that $p\left(\vz|\vx\right)=p\left(\vz\right)$, a sensible choice of prior would be a uniform distribution across all possible labels. Choosing $[\valpha_{\vx}]_{k} = [\valpha]_{k} = \frac{\beta}{k} + 1$ for all $k \in \{1,\dots,K\}$ for some hyper-parameter $\beta$, the MAP objective becomes
    \begin{align}
        \mathcal{L}_{LS} = \frac{1}{N}\sum_{n=1}^N -\log [\vz]_{y_n} + \frac{\beta}{N}\sum_{n=1}^N \sum^K_{k=1} -\frac{1}{K} \log [\vz]_{k}.
        \label{eq:LS}
    \end{align}
    
    With some simple rearrangement of terms, 
    \begin{align*}
        \mathcal{L}_{LS} &= \frac{1}{N}\sum_{i=1}^N -\log [\vz]_{y_n} + \frac{\beta}{N}\sum_{n=1}^N \sum^K_{k=1} -\frac{1}{K} \log [\vz]_{k} \\
        % &= -\sum_{i=1}^n \left( \left(1 + \frac{\beta}{k}\right)\log [\vz]_{y_i} + \sum_{c\neq y_i} \frac{\beta}{k} \log [\vz]_{c} \right) \\
        &= -\frac{\left( 1+\beta \right)}{N}\sum_{n=1}^N \left(\frac{K+\beta}{K(1+\beta)} \log [\vz]_{y_n} + \sum_{K\neq y_n} \frac{\beta}{K(1+\beta)} \log [\vz]_{K} \right)
    \end{align*}
    Thus the above objective is equivalent to the label smoothing regularization with $1 - \epsilon = \frac{k+\beta}{k(1+\beta)}$, up to a constant factor of $(1+\beta)$.
\end{proof}

%%%%%%%%%%%%%%%%%%%%%%%%%%%%%%%%%%%%%%%%%%%%%%%%%%%%%%%%%%%%%%%%%%%%%%%%%%%%%%%
%%%%%%%%%%%%%%%%%%%%%%%%%%%%%%%%%%%%%%%%%%%%%%%%%%%%%%%%%%%%%%%%%%%%%%%%%%%%%%%

\subsection{Proof of \cref{lemma:norm}}\label{subsec:norm_proof}

\textbf{\cref{lemma:norm}}. Let \(\mC \in \mathbb{R}^{D \times \lrbracknorm{\mV}}\) be a matrix with \(\lrbrackvec{\mC}_2 = \sigma_C\), and let \(\vh \in \mathbb{R}^D\) be a vector such that each entry of \(\vh\) satisfies \(|h_i| \leq \sigma_h\). The upper bound for \(\lrbrackvec{\mC\vh}_2\) is:
    \[
    \lrbrackvec{\mC\vh}_2 \leq \sigma_C \cdot \sigma_h \cdot \sqrt{D} .
    \]
\begin{proof}
For any vector \(\vh \in \mathbb{R}^D\), it follows that:
\[
\lrbrackvec{\mC\vh}_2 \leq \lrbrackvec{\mC}_2 \cdot \left\|\vh\right\|_2.
\]
Substituting \(\lrbrackvec{\mC}_2 = \sigma_C\), we obtain:
\[
\lrbrackvec{\mC\vh}_2 \leq \sigma_C \cdot \left\|\vh\right\|_2.
\]
And 
\[
\left\|\vh\right\|_2 \leq \sqrt{\sum_{i=1}^D \sigma_h^2} = \sqrt{D} \cdot \sigma_h.
\]
Substituting the bound on \(\left\|\vh\right\|_2\) into the inequality for \(\lrbrackvec{\mC\vh}_2\), we have the norm of logit vector $u\in \R^V$:
\[\|\vu\|_2 =
\lrbrackvec{\mC\vh}_2 \leq \sigma_C \cdot \left\|\vh\right\|_2 \leq \sigma_C \cdot \sqrt{D} \cdot \sigma_h.
\]
\end{proof}

\subsection{Proof of \cref{thm:entropy_bound}}
\begin{proof}
We begin our analysis of the entropy of the prediction distribution of LM head $\vp \in \R^V$.
\begin{align}
    p_i = \frac{\exp(u_i)}{\sum_{j=1}^V\exp(u_j)}.
\end{align}
The entropy of $\vp$ is then:
\begin{align}
    \mathcal{H}(\vp) = -\sum_{j=1}^Vp_i\log(p_i).
\end{align}
    Without loss of generality, we assume $\|\vu\|_2 =
 \leq \rho \sqrt{ D}$, where $\rho=\sigma_c \sigma_h$. Then we aim to address the following constrained optimization problem:
    \begin{align}
        \min_{\vu}\mathcal{H}(\vu),\, s.t \, \, \|\vu \| \leq \rho \cdot \sqrt{D},
    \end{align}
We derive the global minimum for it with a Lagrangian multiplier and set the corresponding gradients equal to 0 then follow the analysis of \citet{entropy_attn}:
\begin{align}
    \mathcal{L}(\vu, \lambda) = \mathcal{H}(\vu) + \lambda(\|\vu\|^2 - \rho^2 D), \\
    \frac{\partial\mathcal{L}(\vu, \lambda)}{\partial u} = 0, \,\, \frac{\partial \mathcal{L}(\vu, \lambda)}{\partial \lambda} = 0.
\end{align}
Then we get:
\begin{align}
    \lambda u_i &= \sum_{j=1}^V\frac{\exp(u_j)}{Z}\left[\delta_{i,j} - \frac{\exp(u_i)}{Z} \right]\left[1 + \log\left(\frac{\exp(u_j)}{Z}\right)\right], \\\nonumber
    &=p_k[\log(p_k) + \mathcal{H}(\vu)].\label{eq:logandentropy}\\
    \|\vu\| &= \rho^2D.
\end{align}
Assume that for the minimizer $u^*$ there exists an index $i$ such that $u_i^* = 0$, we have:
\begin{align}
    \log(p_i^*) + \mathcal{H}(\vu) = -\sum_{j=1}^{\left|\mV\right|}p_j\log\left(\frac{p_j}{p_i^*}\right) = -\sum_{j=1}^{\left|\mV\right|}p_j\log(e^{u_j})= -\mathbb{E}u.
\end{align}
\begin{align}
    \forall u_{m} \neq 0, u_{n} \neq 0, \,\, p_m\frac{\log(p_m)+\mathcal{H}(\vu)}{u_m} = p_n\frac{\log(p_n)+\mathcal{H}(\vu)}{u_n} \\\nonumber
    \longrightarrow p_m + \frac{\mathbb{E}u}{u_m} = p_n + \frac{\mathbb{E}u}{u_n}
    \longleftrightarrow p_m = p_n.
\end{align}
which contradict to $\|\vu\| = \rho^2D$.
Instead, assume $\forall_n u_n \neq 0$, based on \cref{eq:logandentropy}, we have:
\begin{align}
    \forall u_n \neq u_m, \,\, \frac{p_m}{u_m}\left[\log(p_m) + \mathcal{H}(\vu)\right] = \frac{p_n}{u_n}\left[\log(p_n) + \mathcal{H}(\vu)\right]\\
    \longrightarrow e^u_m\left(1-\frac{\mathbb{E}\left(u\right)}{u_m}\right) = e^u_n\left(1-\frac{\mathbb{E}\left(u\right)}{u_n}\right) \label{eq:equal_func}.
\end{align}
We could assume that a solution \( \vu \) must contain at least one negative component. To illustrate this, consider \( \vu \) where \( u_i > 0 \) component-wise and \( \|\vu\| \leq \rho \sqrt{D} \). We can always shift \( \vu \) by a vector \( \vv \), where \( v_m = v_n \) for all \( m,n \), ensuring that \( \|\vu - \vv\| \leq \rho \sqrt{D} \) and that \( \vu - \vv \) has at least one negative component. Since all components of \( \vv \) are equal and softmax is shift-invariant, it follows that \( \mathrm{softmax}(\vu) = \mathrm{softmax}(\vu -\vv) \). Additionally, without loss of generality, we assume \( \mathbb{E}u > 0 \) based on the same reasoning.
Let $u_m, u_n < 0$, then based on \cref{eq:equal_func}:
\begin{align}
    e^u_m\left(1-\frac{\mathbb{E}(u)}{u_m}\right) = e^u_n\left(1-\frac{\mathbb{E}(u)}{u_n}\right) > 0
\end{align}
As $f(x) = e^x\left(1-\frac{\alpha}{x}\right)$ is monotonously increasing in $x \in (-\infty, 0)$ and $x \in [\alpha, \infty)$ for $\alpha > 0$, it is easy to see $u_m = u_n$. If $u_n < 0$ and $u_m >0$, then $u_m > \mathbb{E}u$. As $f(x)$ is monotonous in $x$ for both $x < 0$ and $x > \alpha$, we have a solution that $\vu$ must have 2 unique values, one positive and one negative. Let the different elements be $a$, $b$. The minimizer $\vu$ contains $\left|\mV\right| - 1$ $b$ and one $a$.
Based on $\|\vu\| = \rho^2D$, we have:
\begin{align}
    a & = \rho\sqrt{D}\sqrt{1-\frac{1}{\left|\mV\right|}}, b = -\rho\sqrt{D}\sqrt{\frac{1}{\left|\mV\right|(\left|\mV\right|-1)}}
\end{align}
The corresponding entropy of the minimizer $\vu$:
\begin{align}
    \log\left(1 + \left(\left|\mV\right|-1\right)\exp\left(-\rho\sqrt{\frac{D\left|\mV\right|}{\left|\mV\right|-1}}\right)\right) + \frac{\rho\sqrt{D\left|\mV\right|\left(\left|\mV\right|-1\right)}\exp\left(-\rho\sqrt{\frac{\left|\mV\right|}{\left|\mV\right|-1}}\right)}{1 + \left(\left|\mV\right|-1\right)\exp\left(-\rho\frac{\left|\mV\right|}{\left|\mV\right|-1}\right)}
\end{align}
\end{proof}

\newpage
\section{Additional Results}\label{app:additional-results}

\subsection{Calibration Metrics}
\begin{figure}[ht!]
    \centering
    \includegraphics[width=0.5\linewidth]{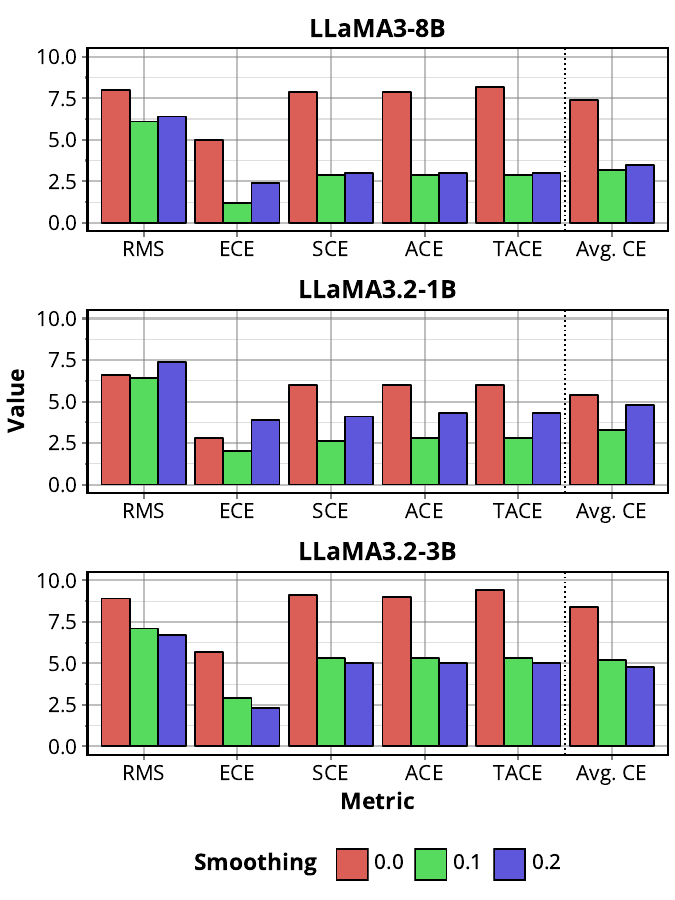}
    \vspace{-0.5cm}
    \caption{Calibration of different \texttt{LLaMA3} models fine-tuned on the same SFT dataset. As the size of the model decreases, the calibration of the model sees less improvement from the use of LS.}
    \label{fig:calibration_llama_tulu3mixture-bars}
\end{figure}

\newpage
\subsection{Reliability Diagrams}
\begin{figure}[h!]
    \centering
    \includegraphics[width=0.7\linewidth]{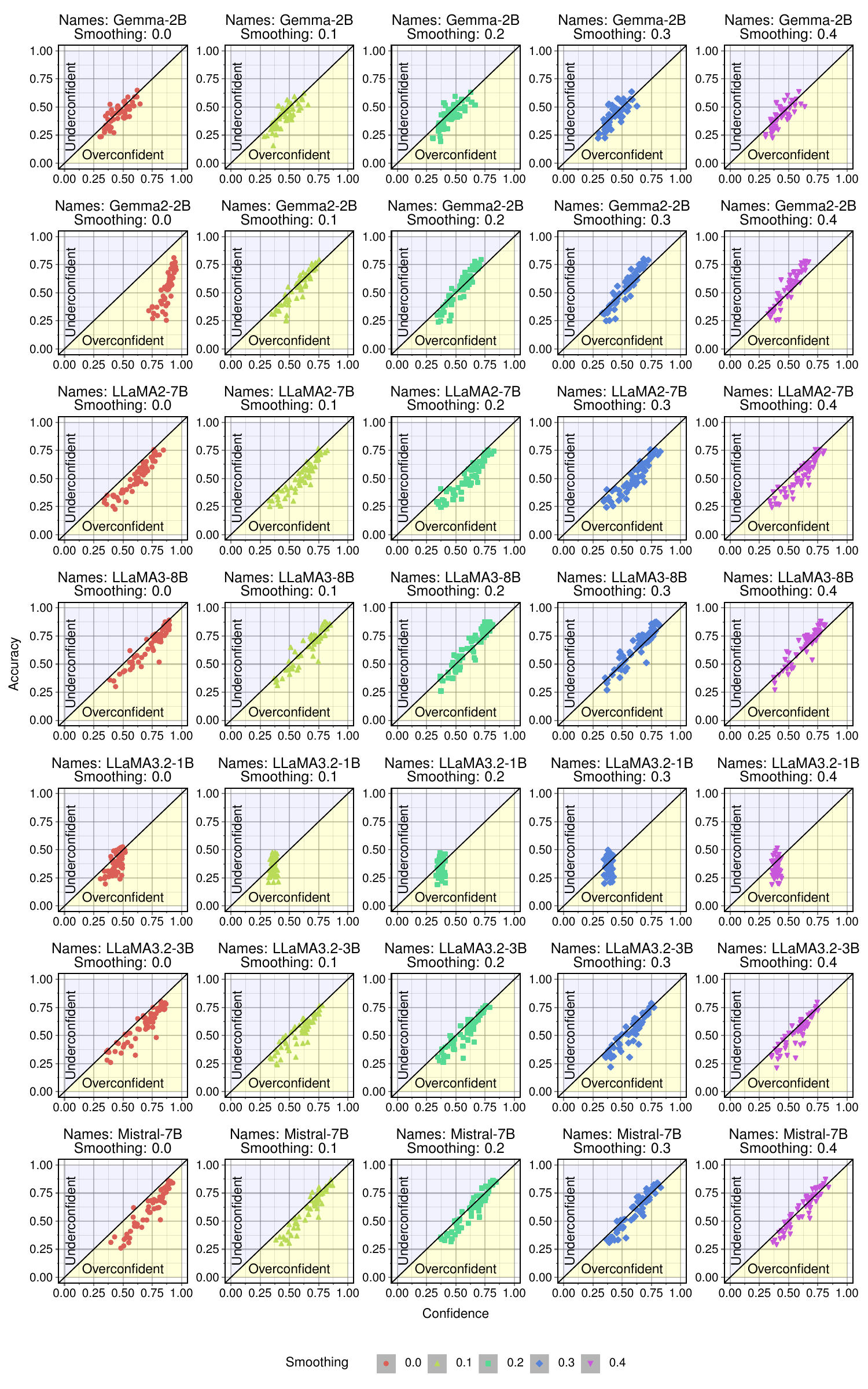}
    \caption{Reliability diagrams of models fine-tuned on the Tulu3 SFT Dataset and tested on the MMLU dataset.}
    \label{fig:reliability_tulu_mmlu}
\end{figure}
\begin{figure}[h!]
    \includegraphics[width=\linewidth]{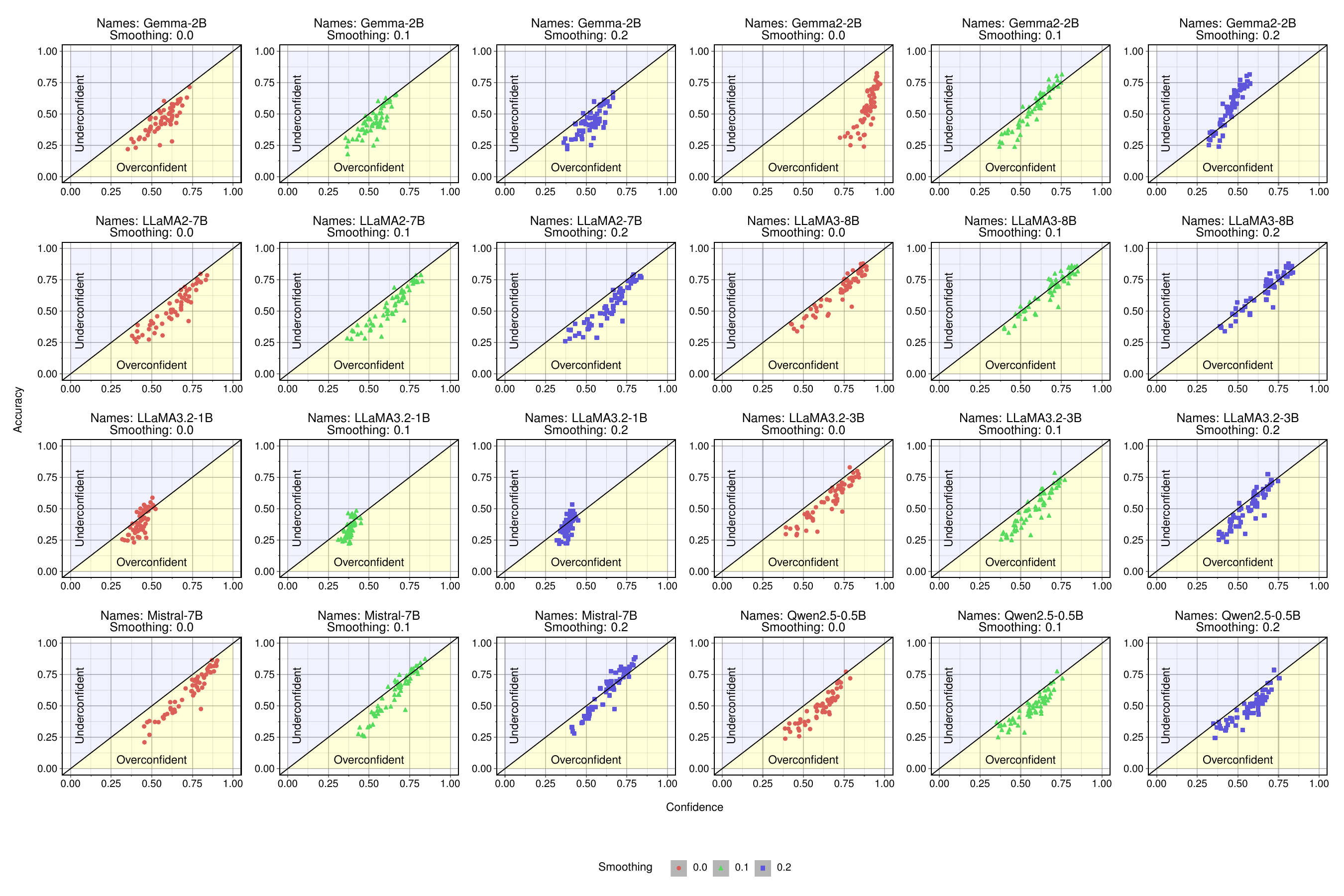}
    \centering
    \caption{Reliability diagrams of models fine-tuned on the OpenHermes-2.5 SFT Dataset and tested on the MMLU dataset.}
    \label{fig:reliability_oh_mmlu}
\end{figure}

\newpage
\subsection{Efficient Smoothed Cross-Entropy}

\subsubsection{Training}

\begin{figure}[h!]
    \centering
    \includegraphics[width=0.8\linewidth]{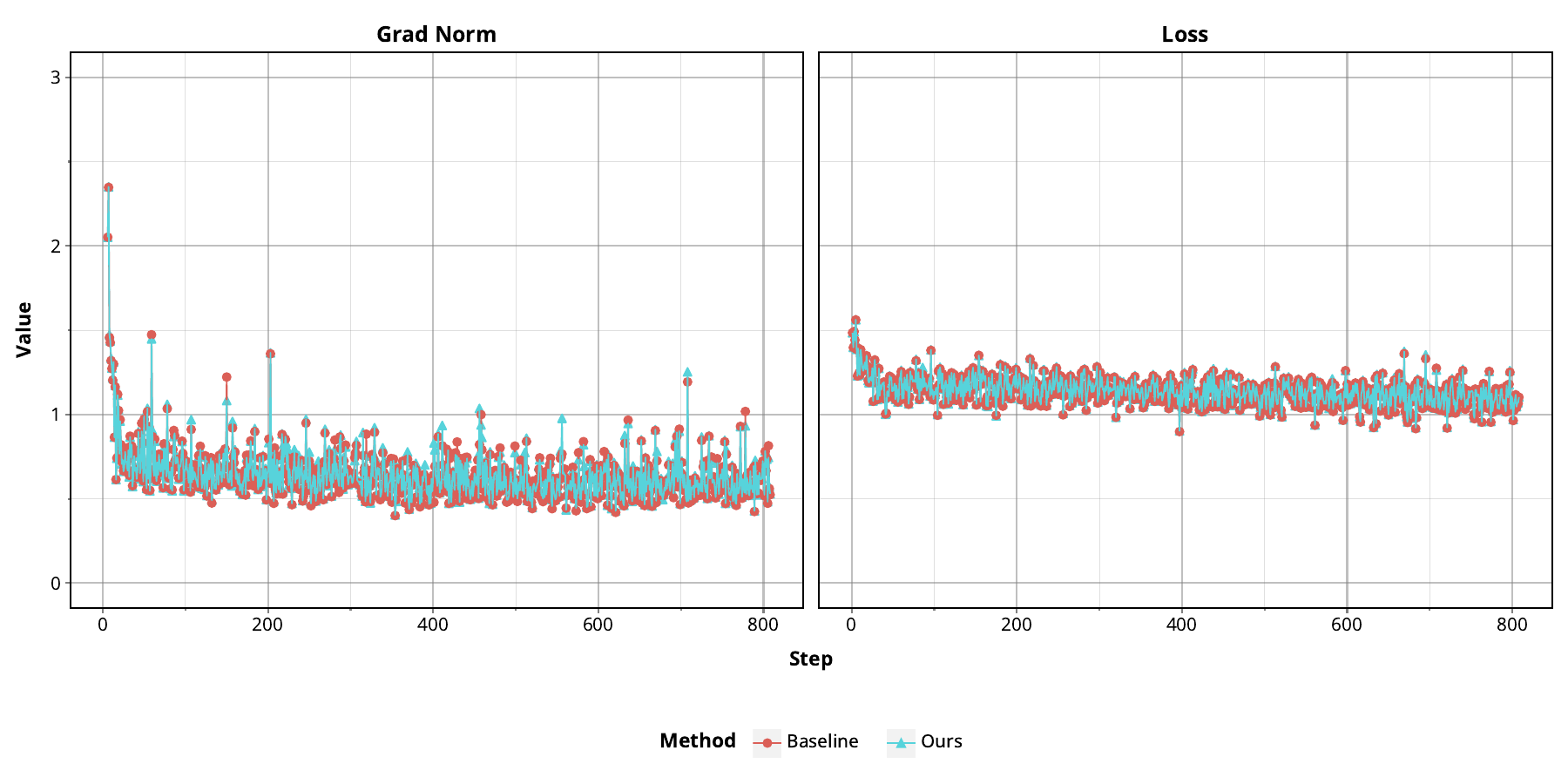}
    \caption{Training curves for \texttt{LLaMA3.2-1B}.}
    \label{fig:enter-label}
\end{figure}
\subsubsection{Benchmarking}
\begin{table*}[h!]
    \centering
    \caption{Memory footprint and time to compute losses and gradients. Results are computed on a batch size of 8192 tokens in a single sequence, generated from a \texttt{Gemma2} with 2 billion parameters (vocabulary size of 256$\mathsf{K}$ and hidden size 2304). Experiments are conducted on an H100-SXM5 GPU with 80GB of RAM, PyTorch 2.4.0 and CUDA 12.1.}
    \label{tab:perf}
    \setlength{\tabcolsep}{2pt}
    \resizebox{\linewidth}{!}{
    \begin{tabular}{
        lc rc rc rc rc rc r}
    \toprule
    \multicolumn{1}{c}{\multirow{2}{*}{\textbf{Method}}} && \multicolumn{3}{c}{\texttt{fwd}} &&  \multicolumn{3}{c}{\texttt{bwd}} && \multicolumn{3}{c}{\texttt{fwd+bwd}} \\
    \cmidrule{3-5}
    \cmidrule{7-9} \cmidrule{11-13}
    && Memory && Time  && Memory  && Time && Memory && Time  \\
    \midrule
    Lower Bound 
        && \qty{0.004}{MB} && 
        && \qty{1161}{MB} &&  
        && \qty{1161}{MB} && \\
    \midrule
    \multicolumn{13}{c}{\textbf{Smoothing $\beta > 0$}} \\
    \midrule
    Ours 
        && \qty{1.1}{MB} && \qty{24.2}{ms}
        && \qty{1163}{MB} && \qty{49.3}{ms} 
        && \qty{1164}{MB} && \qty{72.9}{ms} \\
    \quad+ (No Vocab Sorting) 
        && \qty{0.09}{MB} && \qty{24.1}{ms} 
        && \qty{1162}{MB} && \qty{62.8}{ms} 
        && \qty{1162}{MB} && \qty{86.8}{ms} \\
    \quad+ (No Grad. Filter) 
        && \qty{0.09}{MB} && \qty{24.0}{ms} 
        && \qty{1162}{MB} && \qty{177.1}{ms} 
        && \qty{1162}{MB} && \qty{201.5}{ms} \\
    Fused Version
        && \qty{0.09}{MB} && \qty{100.5}{ms} 
        && \qty{1162}{MB} && \qty{0.2}{ms} 
        && \qty{1162}{MB} && \qty{100.6}{ms} \\
    \quad+ (No Grad. Filter) 
        && \qty{0.09}{MB} && \qty{200.5}{ms} 
        && \qty{1162}{MB} && \qty{0.2}{ms} 
        && \qty{1162}{MB} && \qty{200.5}{ms} \\
    \midrule
    Liger Kernels~\citep{liger}\tablefootnote{The gradient and loss are computed simultaneously, not in separate forward/backward passes.}
        && \qty{}{NA} && \qty{}{NA} 
        && \qty{}{NA} && \qty{}{NA} 
        && \qty{5349}{MB} && \qty{155.0}{ms} \\
    % \texttt{torchtune}~\citep{torchtune} (8 chunks) 
    %     && \qty{8000}{MB} && \qty{}{ms} 
    %     && \qty{1630}{MB} && \qty{}{ms} 
    %     && \qty{9631}{MB} && \qty{}{ms} \\
    \texttt{torch.compile} 
        && \qty{4000}{MB} && \qty{22.8}{ms} 
        && \qty{12000}{MB} && \qty{38.3}{ms} 
        && \qty{16000}{MB} && \qty{62.3}{ms} \\
    Baseline (\texttt{torch.nn.CrossEntropyLoss}) 
        && \qty{24000}{MB} && \qty{41.4}{ms} 
        && \qty{16000}{MB} && \qty{62.5}{ms} 
        && \qty{28000}{MB} && \qty{104.9}{ms} \\
    \midrule
    \multicolumn{13}{c}{\textbf{Smoothing $\beta = 0$}} \\
    \midrule
    Ours 
        && \qty{1.1}{MB} && \qty{24.0}{ms} 
        && \qty{1163}{MB} && \qty{49.2}{ms} 
        && \qty{1164}{MB} && \qty{72.9}{ms}\\
    \quad+ (No Vocab Sorting) 
        && \qty{0.09}{MB} && \qty{23.9}{ms} 
        && \qty{1162}{MB} && \qty{62.4}{ms} 
        && \qty{1162}{MB} && \qty{85.2}{ms} \\
    \quad+ (No Grad. Filter) 
        && \qty{0.09}{MB} && \qty{23.9}{ms} 
        && \qty{1162}{MB} && \qty{177.3}{ms} 
        && \qty{1162}{MB} && \qty{201.4}{ms} \\
    Fused Version
        && \qty{0.09}{MB} && \qty{100.5}{ms} 
        && \qty{1162}{MB} && \qty{0.2}{ms} 
        && \qty{1162}{MB} && \qty{100.6}{ms} \\
    \quad+ (No Grad. Filter) 
        && \qty{0.09}{MB} && \qty{200.5}{ms} 
        && \qty{1162}{MB} && \qty{0.2}{ms} 
        && \qty{1162}{MB} && \qty{200.5}{ms} \\
    \midrule
    Cut-Cross Entropy~\citep{CCE_loss}\tablefootnote{This method does not support label smoothing.} 
        && \qty{1.1}{MB} && \qty{23.6}{ms}
        && \qty{1163}{MB} && \qty{49.2}{ms} 
        && \qty{1164}{MB} && \qty{72.4}{ms} \\
    \quad+ (No Vocab Sorting) 
        && \qty{0.09}{MB} && \qty{23.5}{ms} 
        && \qty{1162}{MB} && \qty{62.5}{ms} 
        && \qty{1162}{MB} && \qty{85.0}{ms} \\
    \quad+ (No Grad. Filter) 
        && \qty{0.09}{MB} && \qty{23.5}{ms} 
        && \qty{1162}{MB} && \qty{177.1}{ms} 
        && \qty{1162}{MB} && \qty{201.4}{ms} \\
    Liger Kernels~\citep{liger}\tablefootnote{The gradient and loss are computed simultaneously, not in separate forward/backward passes.}
        && \qty{}{NA} && \qty{}{NA} 
        && \qty{}{NA} && \qty{}{NA} 
        && \qty{5349}{MB} && \qty{154.8}{ms} \\
    Chunked Cross Entropy (\texttt{torchtune})~\citep{torchtune} (8 chunks) 
        && \qty{13000}{MB} && \qty{30.4}{ms} 
        && \qty{2000}{MB} && \qty{51.0}{ms} 
        && \qty{13000}{MB} && \qty{82.8}{ms} \\
    \texttt{torch.compile}
        && \qty{4000}{MB} && \qty{20.6}{ms} 
        && \qty{4000}{MB} && \qty{33.9}{ms}
        && \qty{8000}{MB} && \qty{55.0}{ms} \\
    Baseline (\texttt{torch.nn.CrossEntropyLoss}) 
        && \qty{24000}{MB} && \qty{38.7}{ms} 
        && \qty{16000}{MB} && \qty{55.8}{ms} 
        && \qty{28000}{MB} && \qty{96.0}{ms} \\
    \bottomrule
    \end{tabular}}
    \gdef\rownumber{}
\end{table*}

\subsection{Experimental Details}
\paragraph{Training. }We conducted a learning rate sweep over learning rates [$\mathrm{5e\text{-}6}, \mathrm{2e\text{-}5}, \mathrm{5e\text{-}5}, \mathrm{2e\text{-}4}$] with a summing reduction. We further tested label smoothing hyper-parameters [$0.0, 0.1, 0.2, 0.3, 0.4, 0.5$], where 0.0 is no smoothing. We used the open-instruct repository at commit \texttt{e363290} for our training setup,\footnote{\url{https://github.com/allenai/open-instruct}} with modifications to account for our kernel as well as specific experimental hyper-parameter settings and baselines.

\paragraph{Evaluation.} Our implementation is based on the MMLU official repository\footnote{\url{https://github.com/hendrycks/test}}. We first evaluate our models on MMLU and then modify the files here to directly adapt the evaluation dataset to the other tasks at hand. We follow MMLU and use the following prompt for all tasks: \textit{`The following are multiple choice questions (with answers) about \{\}.{\textbackslash n}{\textbackslash n}'.format(query)}.
\end{document}